\documentclass[runningheads]{llncs}
\usepackage{graphicx}
\usepackage{times}
\usepackage{latexsym}
\usepackage{wrapfig,lipsum,booktabs}
\usepackage{algorithm}
\usepackage[table,xcdraw]{xcolor}
\usepackage{algpseudocode}
\RequirePackage{algorithm}
\usepackage{amsmath}
\usepackage{color}
\usepackage{caption}
\RequirePackage{amsmath}
\usepackage{graphicx}
\usepackage{colortbl}
\usepackage{array}
\usepackage[hidelinks]{hyperref}
\RequirePackage{bm}
\usepackage{multirow}
\usepackage{arydshln}
\usepackage{paralist}
\usepackage{longtable}
\usepackage[para]{footmisc}
\usepackage[table]{xcolor}
\usepackage{xcolor, soul}
\usepackage[framemethod=default]{mdframed}

\usepackage{soul}
\usepackage{tikz}
\usetikzlibrary{calc}
\usetikzlibrary{decorations.pathmorphing}

\definecolor{etonblue}{rgb}{0.59, 0.78, 0.64}
\definecolor{lightblue}{rgb}{0.68, 0.85, 0.9}
\definecolor{lightgreen}{rgb}{0.56, 0.93, 0.56}

\usepackage[T1]{fontenc}
\newcommand{\am}[1]{\textcolor{red}{#1 -- AM}}
\newcommand{\SB}[1]{\textcolor{blue}{#1 -- SB}}
\newcommand{\rh}[1]{\textcolor{brown}{#1 -- RH}}

\begin{document}
%
%\title{Contribution Title\thanks{Supported by organization x.}}
\title{\textsc{DistALANER}: Distantly Supervised Active Learning Augmented Named Entity Recognition in the\\ Open Source Software Ecosystem}
\author{Somnath Banerjee\inst{1}\and
Avik Dutta\inst{1}\and
Aaditya Agrawal\inst{1}\and
Rima Hazra\inst{1,2}\and \\
Animesh Mukherjee\inst{1}}

\authorrunning{Banerjee et al.}
\titlerunning{DistALANER}% Part of RIGHT running header
\institute{Indian Institute of Technology Kharagpur, India\\
\and Singapore University of Technology and Design\\
\email{\{som.iitkgpcse\}@kgpian.iitkgp.ac.in, \{rima\_hazra\}@sutd.edu.sg}}
\titlerunning{\textsc{DistALANER}}
\maketitle            % typeset the header of the contribution
\begin{abstract}

As the AI revolution unfolds, the push toward automating support systems in diverse professional fields ranging from open-source software to healthcare, and banking to transportation has become more pronounced. Central to the automation of these systems is the early detection of named entities, a task that is foundational yet fraught with challenges due to the need for domain-specific expert annotations amid a backdrop of specialized terminologies, making the process both costly and complex. In response to this challenge, our paper presents an innovative named entity recognition (NER) framework~\footnote{\url{https://github.com/NeuralSentinel/DistALANER}} tailored for the open-source software domain. Our method stands out by employing a distantly supervised, two-step annotation process that cleverly exploits language heuristics, bespoke lookup tables, external knowledge bases, and an active learning model. This multifaceted strategy not only elevates model performance but also addresses the critical hurdles of high costs and the dearth of expert annotators. A notable achievement of our approach is its capability to enable pre-large language models (pre-LLMs) to significantly outperform specially designed generic/domain specific LLMs for NER tasks. We also show the effectiveness of NER in the downstream task of relation extraction. 

%To ensure transparency and facilitate future research in this domain, we are committed to making all our codes, datasets, and models openly available in the public domain upon acceptance\footnote{https://anonymous.4open.science/r/DistALANER-DC9D/README.md}

\end{abstract}
\section{Introduction}
%Named entity recognition (NER) is a sub-task of information extraction that seeks to locate and classify named entities in text into predefined categories such as person names, organizations, locations etc. NER-related tasks can include other aspects of information extraction like co-reference resolution~\cite{dobrovolskii-2021-word, kirstain-etal-2021-coreference}, relation extraction~\cite{huguet-cabot-navigli-2021-rebel-relation}, and event extraction~\cite{huang-jia-2021-exploring-sentence}. 
%,Krallinger2015,nedellec-etal-2013-overview
Traditional named entity recognition (NER) models exhibit certain limitations, particularly when dealing with domain-specific data. Primarily, this stems from the fact that NER models are conventionally trained on generic corpora, rendering them less effective when encountering text sourced from specialized domains such as software, legal~\cite{trias-etal-2021-named}, biomedical~\cite{Jensen2006-ef} or engineering fields~\cite{Li_2022}, which inherently possess distinctive vocabularies and entities. %For instance, ``java'' could denote a programming language in a computational context, but refer to a geographical island in another.
For instance, the word ``\textit{windows}'' could denote a well-known operating system in the realm of software and technology, yet simultaneously refer to a commonplace architectural feature in the context of residential or commercial structures. Adapting a generic NER model to handle tasks specific to a certain field often requires extra, specialized training data from that area. However, gathering and annotating this additional data can be a costly and time-consuming task.
% \vspace*{-0.1cm}
% \begin{figure}[htb]%[!ht]
% \centering
% \vspace*{-0.3cm}
%   \includegraphics[width=0.7\linewidth]{Figures/DistantvsHuman.pdf}
%     \caption{\footnotesize {Annotation we obtain from \textsc{DistALANER} vs ground truth.}}
%   \label{fig:entityAll}
%   \vspace*{-0.25cm}
% \end{figure}\\
%,fang-etal-2021-tebner
\noindent Distant supervision~\cite{Liang:2020} methods help solve the problem of insufficient labels by automatically generating labeled data for entity recognition. Using a raw text and a dictionary, these methods label entities through exact string matching, then use this data to train advanced neural models for recognizing entities. However, two key challenges arise from this approach. The first is incomplete annotations~\cite{jie-etal-2019-better}. Many dictionaries do not fully cover domain-specific entities, leading to a lot of unmatched entities and false-negative labels. Earlier attempts to increase labeled entities involved expanding the dictionary with set rules~\cite{liu2019hamner}, but these rules are often hard to apply in other fields. The second challenge is the struggle to identify new, unannotated entities. Even models that are manually trained have difficulty with this due to their limited capabilities.
\begin{wrapfigure}{r}{0.50\textwidth} %this figure will be at the right
    \centering
    \vspace*{-0.5cm}
    \includegraphics[width=0.50\textwidth]{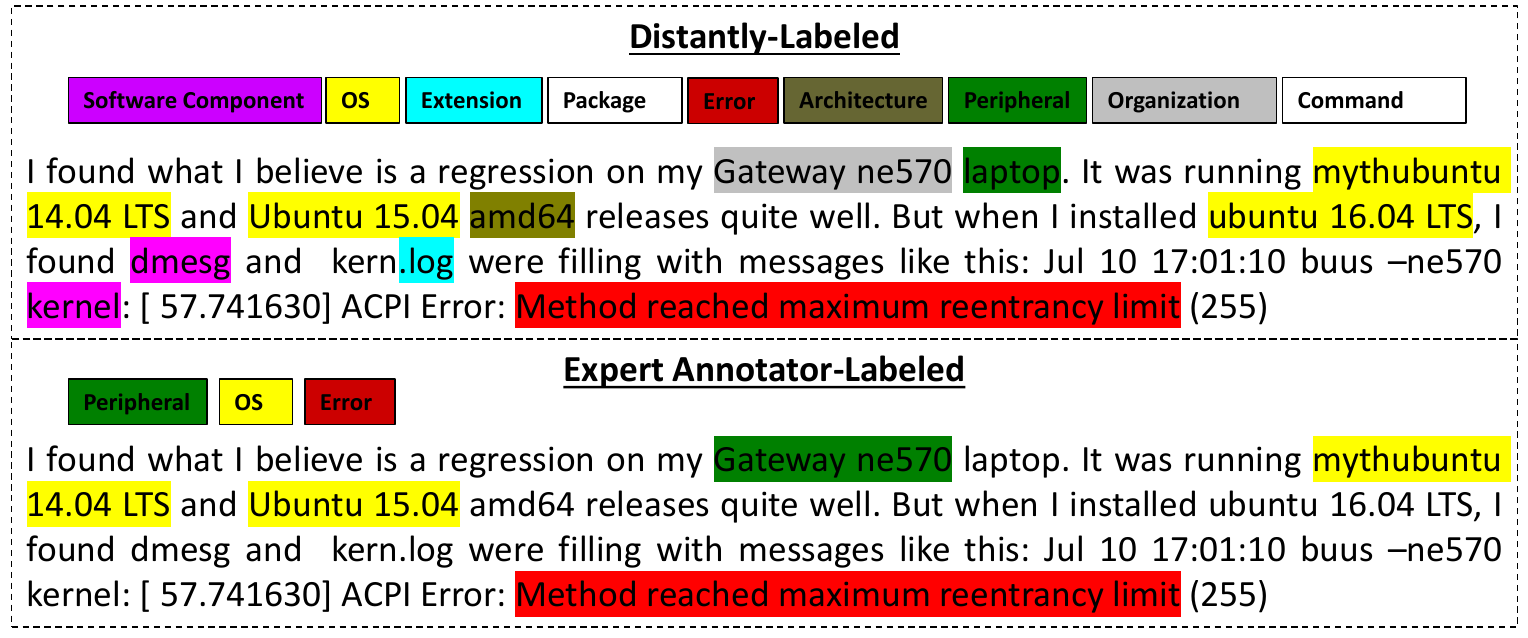}
    \caption{\footnotesize {Annotation we obtain from \textsc{DistALANER} vs ground truth.}}
      \label{fig:entityAll}
  \vspace*{-0.25cm}
\end{wrapfigure}
\noindent In the area of open source software development, the need for NER has become increasingly critical~\cite{Tang2023}. NER plays a pivotal role in deciphering and categorizing textual information into predefined entities such as individual contributors, programming languages, software tools, and project specifications found in software documentation, source code, bug reports, or community discussions~\cite{wang-etal-2022-named}. The understanding and categorization of such entities offer deep insights, allowing for effective communication, resource allocation, and decision-making within the open source community. %In addition, it assists in knowledge extraction~\cite{10.3389/fcell.2020.00673}, facilitating a greater comprehension of the system's functionality~\cite{shrimal-etal-2022-ner}, feature requirements, software dependencies, or potential bugs~\cite{ZHOU2020110572}, thereby augmenting the efficiency of software development and maintenance. 
Further, NER can aid in community management tasks, such as identifying contributors and their areas of expertise or mapping the interactions within the developer community. Consequently, the integration of NER in the open source software domain could dramatically streamline processes, enhance collaboration, and eventually improve the overall quality of the software produced.\\
\noindent Recently introduced LLMs can be highly effective in the software domain~\cite{10109345}. With their ability to understand complex patterns and generate human-like text, they can assist in identifying and classifying key entities such as specific coding languages, software tools, packages, peripherals, or developers mentioned in various sources like code, software documentation, and community discussions. However, LLMs can sometimes prove to be a bottleneck in identifying NER due to security issues, cost and lack of contextual knowledge. These limitations arise due to their broad contextual learning from vast corpora that often spans numerous domains, making them less specialized for a particular field like software development. These models might struggle to identify and classify domain-specific entities accurately, given their generic training~\cite{qin2023chatgpt}. Moreover, as LLMs learn from data available up to their last training checkpoint; they may not be aware of new terms or entities introduced in the domain post-training. Taking into account the aforementioned limitations~\cite{yang2023harnessing}, we put forth an innovative framework, explicitly designed for the software domain trained through bug data, manuals and CQAs.
%Our aim is to use lightweight models in our framework so they easily serve as a preliminary stage in larger systems. We require minimal human effort in this framework. 
Our objective is to incorporate lightweight models into our framework, allowing them to seamlessly integrate with any system. Our approach involves limited human intervention. By including a variety of methods depicted in Figure~\ref{fig2}, we enhance the efficiency of deep learning models. The key contributions of this paper are as follows. 
%Here, we are not focusing towards proposing any complex architecture based heavy methods (some of them existed in literature already). Our novelty lies in experimenting on light weight older basic models with distantly annotated data and make it working to easily adaptable in larger system.
%This aims to augment the proficiency of deep learning models by incorporating a diverse set of methodologies as outlined in Figure~\ref{fig2}. %such as ``Software Domain Heuristics'', ``Time Series Software Components'', ``External Knowledge on Software Entities'', a ``Unique Logbook Derived from Open Source Ecosystem Manuals'', ``POS Corrector'' (see Figure~\ref{fig:dupTrn}) and an ``Active Learning Strategy for Knowledge Updates''. 
%The key contributions of this paper are as follows.
\begin{compactitem}
    \item In terms of datasets we release the following items - (a) a large open source domain corpus with both human and system annotations for nine different named entity types (see examples in Figure~\ref{fig:entityAll}), (b) For each entity type, a large unique lookup table contains relevant entities automatically collected from various sources since 2004 (to check detailed sources, see section~\ref{sec:sourcedetail}) and
    %a huge unique lookup table consisting of all open source domain package, command, new features~\SB{New feature?-- Can we write this? -- For each entity type, huge unique lookup table has been constructed by automatically collected relevant entities from various sources (to check detailed sources, see appendix). } and external knowledge from the year 2004 and 
    (c) a large corpus of human annotated entity relation pairs in software domain for the downstream application task.%\footnote{https://zenodo.org/record/8075578}. 
    \item We put forth an innovative method for expanding the dictionary with the largest software domain specific data that contains rich temporal context and is not reliant on either ambiguous strings or ad hoc rules. Experimental results confirm that our method markedly enhances the quality of annotations produced through distant supervision.
    \item We conduct extensive experiments on four large datasets with \textsc{DistALANER} framework achieving the best performance with minimal human efforts. Utilising our framework in conjunction with pre-LLM era models outperforms LLMs like GPT-3.5-Turbo\footnote{https://openai.com/blog/chatgpt}, GPT-4\footnote{https://platform.openai.com/playground}, Google-BARD\footnote{https://bard.google.com/} and task specialized UniversalNER\footnote{https://universal-ner.github.io/} by a substantial margin.
    %\item In order to demonstrate the effectiveness of the named entities extracted by our method we use it to perform the closely associated problem of relation extraction reporting superior performance.~\SB{I feel we can remove it!}
\end{compactitem}

\section{Related work}
\noindent\textit{Named entity recognition in software ecosystems}: The landscape of Named Entity Recognition (NER) in software ecosystems has evolved significantly, incorporating diverse approaches and datasets. \cite{Ye:2016} pioneered S-NER to identify software-related entities like programming languages and APIs using a subset of the 2015 StackOverflow dataset. This effort was complemented by manual annotations for supervised learning. A shift towards a unified framework was made by \cite{li-etal-2020-unified,ZHOU2020110572}, employing machine reading comprehension to identify entities via answer spans rather than sequence labeling, tested on both nested and flat NER datasets. In parallel, \cite{tabassum-etal-2020-code} introduced SoftNER, leveraging a programming-related StackOverflow dataset and BERT \cite{devlin2019bert} for enhanced recognition of code tokens and software entities. Meanwhile, \cite{kocaman2020biomedical} showcased BiLSTM-CNN-Char for biomedical entity extraction, emphasizing efficiency without heavy transformer reliance. The field continues to advance with the introduction of the FEW-NERD dataset for few-shot, multigrained, and nested NER \cite{ding-etal-2021-nerd}, alongside \cite{xia-etal-2019-multi}'s MGNER framework for multi-granularity entity detection, highlighting the ongoing innovation in entity recognition technologies.

\noindent\textit{Distant supervision based named entity recognition}: In~\cite{fang-etal-2021-tebner,peng-etal-2019-distantly}, authors introduced innovative methods to enhance distantly supervised Named Entity Recognition (NER) through dictionary extension, type expansion, and leveraging distantly labeled data for model training. These approaches aimed to automate data labeling and entity identification, demonstrating superior performance over existing systems. Additionally, \cite{Liang:2020} and~\cite{wang2023gptner} proposed the BOND framework and GPT-NER respectively, utilizing pretrained language models and a generative framework to further improve NER model efficacy.
\section{Dataset}
\label{sec:dataset}
%\SB{As mentioned in EACL do we need to move the dataset and source details after methodology? Sir's opinion!} \am{No}
%\begin{table}[!h]
\begin{wraptable}{l}{5.5cm}
\vspace*{-0.75cm}
\centering
%\small
\scalebox{0.78}{
\begin{tabular}{|l|c|}
\hline
\textbf{Properties}                    & \multicolumn{1}{l|}{\textbf{Ubuntu bug count}} \\ \hline
\#bugs before filtering              & 270K                                           \\ \hline
\#bugs after filtering              & 170K                                           \\ \hline
Avg \#words in description & 141                                            \\ \hline
Max \#words in description & 399                                            \\ \hline
Min \#words in description & 60                                             \\ \hline
\end{tabular}
}
\caption{\footnotesize The basic statistics of the Ubuntu bug dataset.}
\label{tab:datastat}
\vspace*{-0.4cm}
%\end{table}
\end{wraptable}
In this paper, we utilize two datasets specifically from the Ubuntu ecosystem: (i) the Ubuntu bug repository and (ii) software community question-answering repositories. The latter is further subdivided into QAs from three different community posts -- Linux, Fedora and Ubuntu thus resulting in a total of four datasets. We use the bug dataset for training the model and the three QA datasets for evaluating its performance. The motivation for selecting bug data for training an open-source NER model are as follows. First, it provides a rich source of diverse and complex natural language text, which includes technical terminology, software components, and descriptions of problems and solutions, making it a well-suited resource for understanding and learning the language structure and context within the open-source domain. Second, bug reports often involve specific named entities such as software component names, version numbers, and user handles, among others, providing a plethora of examples for NER tasks. Further, the nature of bug tracking in open-source projects often involves collaboration and communication between various contributors, yielding a wide variety of linguistic styles and expressions. This diversity enhances the model's robustness and adaptability. Last, as bug data is openly accessible, it aligns with the open-source ethos of shared knowledge, making it an appropriate dataset for open-source NER model training.
\noindent \textbf{Ubuntu bug repository}: In our experiment, we use the repository of bugs collected by~\cite{Hazra:2021}. These bugs are mainly reported on packages, conflicts between Ubuntu and Windows and other related events. The dataset contains approximately 270K bugs along with the metadata such as title of the bug, description of the bug, user who posted the bug, comments and their commenters, creation date of the bug, tags of the bugs. In our work, we mainly use the description of the bugs to obtain the named entities. We filter this raw dataset by excluding (i) those bugs that solely reference another bug, for example, ``Automatically imported from Debian bug report \#257568"\footnote{http://bugs.debian.org/257568} and (ii) those bugs that have very small description size ($<60$ words) or exceedingly large description size ($>400$ words). The bugs with very small description size prohibits obtaining meaningful representations while those with very large size routinely include code logs in large proportions rather than useful text. The dataset statistics are presented in Table~\ref{tab:datastat}.
\noindent \textbf{QA datasets}: We choose Ubuntu\footnote{https://launchpad.net/ubuntu}, Fedora \footnote{https://forums.fedoraforum.org/} and Linux\footnote{https://www.linux.org/forums/} question-answering community posts for the purpose of evaluation. These posts contain questions on the respective open-source system and the many problems related to it. Each such question has a title, a body, an asker identity, a posting date and time, a set of answers, the answerer identity and the answer posting time. For our purpose, we annotate a total of 500 question-answer pairs from each community. These question-answer pairs are chosen randomly to ensure an unbiased sampling. %We attempt to select similar datasets which belong to same ecosystem.
% ~\rh{To Somnath: Feedback:\\
% 1. Can't we reduce the volume of reasons for using Bug data?
% 2. Dataset table only contains bug data. If we are not going to present the launchpad data then do not refer the table in the launchpad data description section.}
% \begin{table}[!h]
% \centering
% \small
% \resizebox{.40\textwidth}{!}{
% \begin{tabular}{|p{3.5cm}| p{2cm}|} \hline
% {\bf Properties} &  {\bf Count} \\ \hline 
% Number of bugs & 270K \\ \hline
% Average number of words in description & \\ \hline
% Maximum number of words in description & \\ \hline
% Minimum number of words in description & \\ \hline
% \end{tabular}
% }
% \caption{\label{tab:dataset} This table contains the basic statistics of Ubuntu bug dataset collected from launchpad.}%\footnotesize
% \end{table}
% \begin{table}[!h]
% \centering
% \small
% \resizebox{.40\textwidth}{!}{
% \begin{tabular}{|p{3.5cm}| p{2cm}|} \hline
% {\bf Properties} &  {\bf Count} \\ \hline 
% Number of questions &  \\ \hline
% Average number of words in question title & \\ \hline
% Maximum number of words in question body & \\ \hline
% Minimum number of words in description & \\ \hline
% \end{tabular}
% }
% \caption{\label{tab:dataset} This table contains the basic statistics of Ubuntu question answering dataset collected from launchpad.}%\footnotesize
% \end{table}

%\vspace*{-0.7cm}
\section{Source details}\label{sec:sourcedetail}
%\SB{Added this whole section from Appendix}
The collection of data for our study comes carefully from various reliable sources. 
\begin{compactitem}
\item \noindent\textit{Operating systems}: Names primarily come from the official Ubuntu pages\footnote{https://wiki.ubuntu.com/Releases} and Wikipedia\footnote{https://en.wikipedia.org/wiki/List\_of\_operating\_systems} for all other operating systems.

\item \noindent\textit{Architecture}: Base architectures come from the community\footnote{https://help.ubuntu.com/community/SupportedArchitectures}, and we manually include additional writing styles.

\item \noindent\textit{Commands}: Commands in structured form come from github\footnote{https://github.com/nengz/ShellFusion}. We collect additional commands using TagMe and add them with our active learning based approach from Wikipedia.

\item \noindent\textit{Packages}: We use all the packages assembled by the authors in~\cite{Hazra:2021}.

\item \noindent\textit{Error codes}: We collect these from the Ubuntu page\footnote{https://wiki.ubuntu.com/error\_and\_warning\_messages}, and use TagMe (with our active learning based approach) to augment the list with additional error codes.

\item \noindent\textit{File extensions}: We collect this data primarily from the Wikipedia page\footnote{https://en.wikipedia.org/wiki/ Filename\_extension}.

\item \noindent\textit{Organizations}: We initially create a base list from a Wikipedia page to capture computer-related organizations, and then use TagMe (with our active learning based approach) to expand the list.

\item \noindent\textit{Peripheral types and software components}: In the absence of a comprehensive list, we prepare an initial handcrafted list. We then use TagMe (with our active learning based approach) to significantly expand the lists of these two entity types.

\end{compactitem}

\section{Preliminaries}
% In a bug repository $\mathcal{B}$, we have a set of bugs denoted by ${b_i}$. Suppose each bug contains a sequence of words ${w_1, w_2\cdots, w_n}$. In our setting, each entity type is denoted by $\emph{E}^i$ where $i$ is the $i^{th}$ entity type. In total, we have nine entity types which are {\bf PKG}, {\bf OS}, {\bf ORG}, {\bf CMD}, {\bf ERR}, {\bf EXT}, {\bf PRP}, {\bf SOC} and {\bf ARC}. The entities are  In our expanded dictionary, we have a set of entities $\{e^i_1, e^i_2, \cdots  e^i_j\}$ where $j$ denotes the number of entities present in the given entity type $\emph{E}^i$. In our setting, the entity could consist of single or multiple words (phrases). So, an entity span can be $\{w_i, w_{i+1}, \cdots w_{j-1}, w_j\}$ where $i$ is the start index and $j$ is the end index and $i< j$. 
%Named entity recognition is a task of identifying important phrases (could be a single word or multiple words) from a given text segment and classify them as a specific entity type. In this section, we briefly describe the distantly supervised NER approach which is different from the traditional supervised method. %Further, we define the named entity recognition tasks. Supervised NER is a technique where each training input sample is associated with the named entity labelled by expert annotators. So, the labels in the supervised NER are less prone to error. Let's denote the labeled dataset as $D_{sup} = {(x_1, y_1), (x_2, y_2), ..., (x_n, y_n)}$, where $x_i$ represents the $i^{th}$ input sample and $y_i$ represents the corresponding named entity labels. 
Distantly supervised NER aims to automatically label the input data by leveraging existing knowledge bases as opposed to the supervised method that depends on gold labels drawn from the training data. The key hypothesis is that if an entity mention appears in the knowledge base and is also present in the unlabelled data then it is likely to be a named entity. We denote the distantly supervised dataset as $D_{dist} = {(x_1, y_1), (x_2, y_2), ..., (x_m, y_m)}$, where $x_i$ represents the $i^\textrm{th}$ input sample and $y_i$ represents the distant labels generated based on heuristics or knowledge bases. While the distantly labelled data could be prone to error, it is very effective in a low resource setting where there is a genuine scarcity of gold labels.\\ %But, in the case of low resource domain specific scenarios, distantly labelled data are helpful to build the model without the intervention of expert annotators.\\
Note that the corpora for our experiments are primarily composed of text drawn from bug repositories and we shall therefore define the notations in terms of this particular corpora. In a bug repository $\mathcal{B}$, each bug is denoted by $b_i$. A bug $b_i$ is a sequence of words $\{w_1, w_2, \cdots, w_n\}$. We denote the entity types as the $\emph{E}_{ename}$ where {\em ename} is the name of the entity type. We define nine entity types as follows -- packages ({\bf PKG}), operating system ({\bf OS}), organization ({\bf ORG}), commands ({\bf CMD}), errors ({\bf ERR}), file extension ({\bf EXT}), peripherals ({\bf PRP}), software components ({\bf SOC}), and architecture ({\bf ARC}). In our setting, an entity may contain single word or multiple words (phrases). The entity span can be defined as $\{w_i, w_{i+1}, \cdots, w_{j-1}, w_j\}$, where $i$ indicates the starting index, $j$ indicates the ending index, and $i \le j$. We use the conventional IO tagging (inside-outside) method. Given a sequence of words  $\{w_1, \cdots, w_i, w_{i+1}, \cdots, w_j, \cdots w_n\}$ we mark as \{${O, \cdots, I_{ename}, I_{ename}, \cdots, I_{ename}, \cdots, O }$\}. An example bug text and its IO tags are shown below. 
\mdfsetup{skipabove=\topskip,skipbelow=\topskip}
\newrobustcmd\ExampleText{%
    
}
\mdfdefinestyle{exampledefault}{%
rightline=true,innerleftmargin=5,innerrightmargin=5,
frametitlerule=true,frametitlerulecolor=black,
frametitlebackgroundcolor=etonblue,
frametitlerulewidth=3pt}
\begin{mdframed}[style=exampledefault]
\scriptsize{
\textbf{Text:} \texttt{After upgrading to \textbf{Ubuntu 18.04} and thus from \textbf{Linux 4.13} to \textbf{Linux 4.15}) the\\ \textbf{Monitor} connected via \textbf{VGA} (through DVI-I) shows a `No Signal' message after\\ amdgpu takes over from efifb and turns off.}\\ \textbf{Entity labels:} \{$O, O, O, I_{OS}, I_{OS}, O, O, O, I_{OS}, I_{OS}, O, I_{OS}, I_{OS}, O, I_{PRP}, O, O, I_{PRP}, O \cdots $\}.
}
%\vspace{-0.5cm}
\end{mdframed}
\begin{figure*}[!ht]
\vspace*{-0.70cm}
\centering
\includegraphics[width=1.0\textwidth]{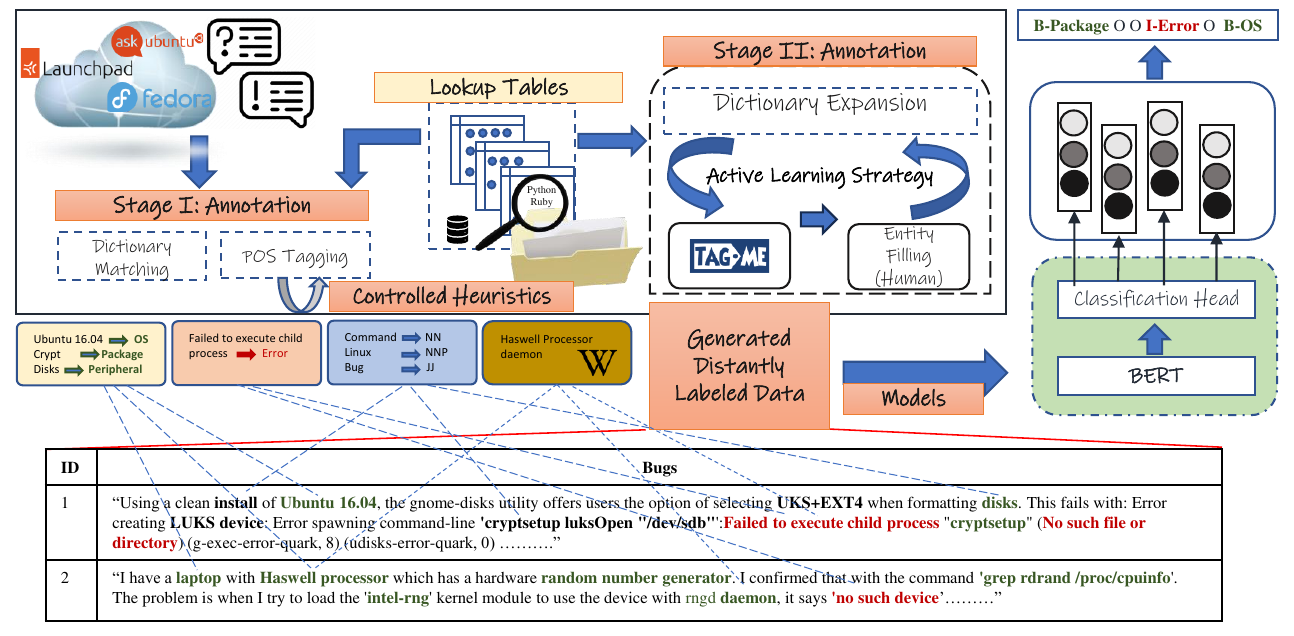}
\caption{\footnotesize Overview of \textsc{DistALANER}. Stage I annotation involves ``dictionary matching'' and ``POS tagging''. Stage II annotation then involves ``dictionary expansion''. After these stages, we identify four types of extractions from data. The light yellow box represents extracted ``entities", while the orange box represents ``error'' types. The blue and dark yellow boxes represent ``POS tags'' and ``Wikipedia mentions'' respectively. We mark these exact types through links for some real bug samples.} %\am{Write one line about each big boxes and the links that you draw.}\SB{updated}}
\label{fig2}
\vspace*{-0.70cm}
\end{figure*}
\section{Methodology}
\vspace*{-0.10cm}
In this section, we introduce the overall framework of \textsc{DistALANER}. We illustrate our framework in Figure~\ref{fig2}. Our framework includes three stages -- (a) Stage 1: Construction and matching of dictionary, (b) Stage 2: Entity distillation and dictionary expansion, (c) Stage 3: Training of the NER model. %Further, we conduct an experiment of application of NER model on relation extraction tasks. Each stage of our framework are described below.
% Annotations in software systems usually require experienced developers. It is quite hard to find out appropriate developers to annotate the entities and the process will be costly. So, we attempt to annotate the data using distant supervision. In this work, we identified a few domain-specific entity types which could play an important role while solving any bugs or answering new questions. We have focused on nine entity types -- Operating system, architecture, command, error code, extension, software organization, package, peripheral and software component. Our framework consists of three components -- (i) Dictionary building. (ii) Dictionary matching, (iii) Dictionary expansion.
\subsection{Stage 1: Construction and matching of dictionary}
At the first stage, we build the dictionary of entities and their respective entity types. To build the dictionary, we use existing knowledge from websites, repositories, and documents of the Ubuntu eco system. %We gather lists of possible entities for the already defined nine entity types from external knowledge bases. There is no centralized knowledge base used to obtain entities. We collect the entities from various sources and build the first-level dictionary. 
\noindent For the \textbf{OS} entity, we include all the Ubuntu distributions (obtained from~\cite{Hazra:2021}) and collect Linux distributions and Windows versions from Wikipedia. For the \textbf{ARC} entity type, we include different writing styles (and versions) of 32 bit and 64 bit \hl{(\texttt{x32} | \texttt{x64}, \texttt{x386} | \texttt{amd64}, etc.)}. For \textbf{CMD} type, we collect all the Linux commands from Wikipedia, Ubuntu man pages. Authors in~\cite{Hazra:2021} collected the list of packages of each Ubuntu distributions (total 20 distributions). For the \textbf{PKG} type, we utilize this list and extract all the unique package names. We build the \textbf{ERR} type by collecting the error codes that usually occur in the Ubuntu system from the Ubuntu wikipage\footnote{https://wiki.ubuntu.com/error/and/warning/messages}. 
%, and the paper~\cite{Zhang:2022}
\begin{wrapfigure}{r}{0.50\textwidth} %this figure will be at the right
    \centering
    \vspace*{-0.7cm}
    \includegraphics[width=6.0cm]{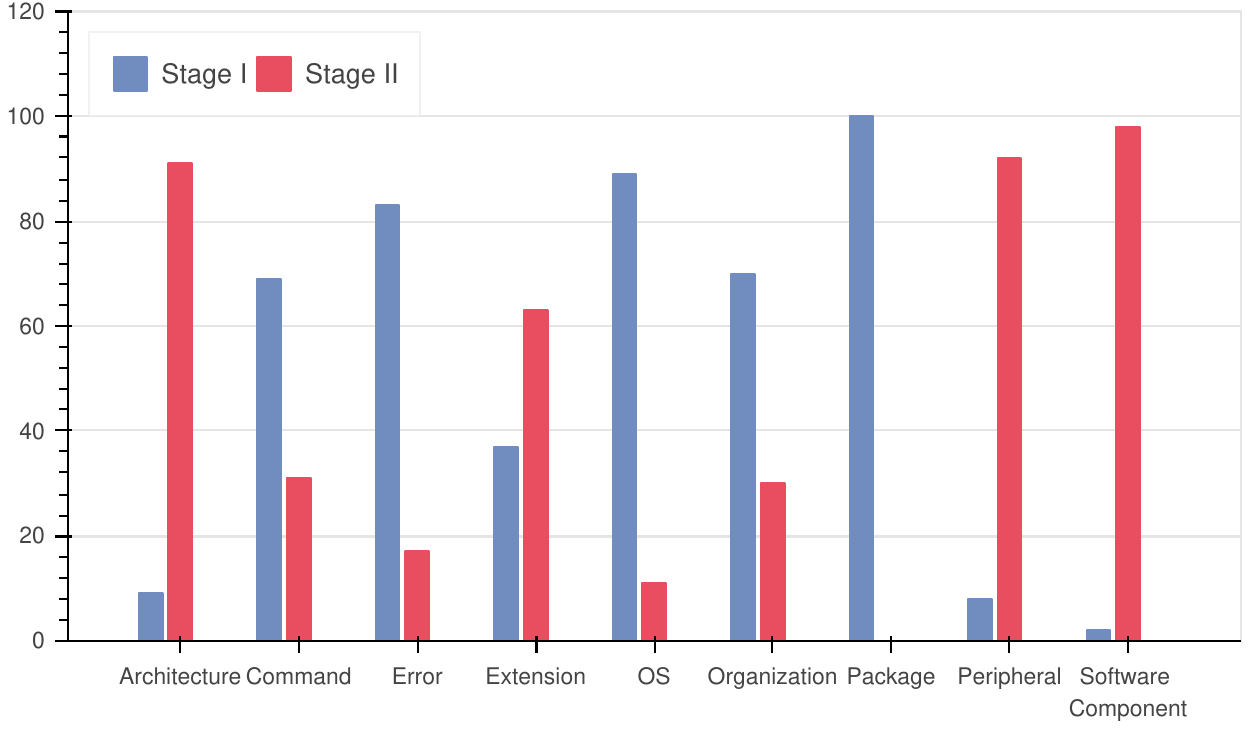}
    \caption{\footnotesize The proportion of entities recognized in Stage I and followed by the proportion extracted from Stage II.}
    \label{fig:galaxy}
    \vspace*{-0.55cm}
\end{wrapfigure}
For the type \textbf{EXT}, we collect data from various Wikipedia pages. For the \textbf{ORG} type, we focus on software organizations and collect the information from Wikipedia. For type \textbf{PRP}, we gather entities from various sources (see section~\ref{sec:sourcedetail}) followed by manual inspection. For \textbf{SOC}, we collect data from various unix-related websites as well as Wikipedia pages (See Table~\ref{tab:entExmpl}). %\am{How do you get SOC?}
%\begin{figure}[!ht]
%\includegraphics[width=1\linewidth]{emnlp2023-latex/Figures/POSFinal.pdf}
%\caption{Samples POS tags excluded based on heuristics. More can be found on Appendix.}
%\label{fig:dupTrn}
%\end{figure}
Next, we conduct dictionary matching. We consider only exact string matching of the bug description text with the entities corresponding to each entity type. For instance, \hl{`\textit{CurrentDesktop}'} is not marked because it contains \hl{`\textit{Desktop}'} as a subword. We discard the wrongly identified entities using a set of regex (see Appendix). %For each entity type, we pass our list of bug descriptions (section~\ref{sec:dataset}) through the implemented regex \SB{for discarding wrongly identified entities} and annotate the phrases accordingly.  
Note that, in earlier paper~\cite{Wang:2020}, researchers have used the \textsc{AutoPhrase} tool~\footnote{https://github.com/shangjingbo1226/ AutoPhrase} to extract phrases from the text, which they later consider for entity identification. We also attempted to apply \textsc{AutoPhrase} to identify entity-containing phrases. However, the tool failed to extract key phrases from our bug descriptions and instead assigns higher scores to irrelevant terms like `\textit{box}', `\textit{release}', `\textit{start}', `\textit{architecture}', `\textit{show}', `\textit{network}', `\textit{subdevice}', and `\textit{software}'. Hence, we resorted to the above dictionary matching technique.
%~\rh{To Somnath: Can we elaborating the dictionary matching part? can we include some steps of doing the matching?}
\if{0}
\noindent\textit{\textbf{RQ: Why AutoPhrase Tool can't be utilized?}}
\noindent \textit{In earlier papers by~\cite{Shang:2018,Wang:2020}, researchers use Autophrase to extract phrases from the text, which they later consider for entity identification. For our task, we apply the Autophrase algorithm to our dataset to try to identify entity-containing phrases. However, it fails to extract key phrases from our bug descriptions, instead assigning higher scores to terms like 'box', 'release', 'start', 'architecture', 'show', 'network', 'subdevice', and 'software'. }\fi

\subsection{Stage 2: Entity distillation and dictionary expansion}
% Our first phase is dictionary matching. For each entity type, we pass our list of bugs through the implemented regex and annotate the phrases accordingly. We consider only exact matching with the entities given an entity type. 
In Stage 2, we distill the exactly matched entities and then expand the list of entities for a given entity type using an active learning approach. Active learning enables our system to learn iteratively, refining its understanding as it receives feedback from its interactions. %It is particularly effective when faced with a large unlabelled dataset, as is the case here. %~\textcolor{red}{As part of the process, after receiving the annotated data from Stage 1 and POS filtered bugs (L) along with their IO mentions, a binary RoBERTa-based classifier is trained. Subsequently, we take up 100*16 = 1600 bugs from TagMe but not included in Stage 1 and feed them into our model for classification as either ``entity” or ``non-entity”. If the confidence for an entity being identified is 50\% or greater, it undergoes manual labeling by a human to categorize it into one of the nine predefined entity types. Once the manual annotation is completed, the newly labeled data (d) is integrated into L, resulting in an updated dataset, (L~$\cup$~d). The model is then retrained using this revised dataset, and the entire procedure is reiterated until the stopping criterion is met. The stopping criterion is defined as the absence of data in the list of entities identified by TagMe.} 
We observe that many phrases get marked as an entity, but in reality, some of them are not entities. Thus, we employ two heuristics to distill the entities: (1) identifying parts-of-speech patterns, and (2) human intervention.
\noindent \textbf{Parts-of-speech patterns}: Here, we find that the likelihood of a phrase being an entity often depends on the parts-of-speech tag of the words in the phrase or before and after the phrase. For instance, the verb `\textit{find}' in a sentence is never an entity, even though our dictionary matching may mistakenly identify it as a \textbf{CMD} entity (see more samples in section~\ref{sec:appendix1}). We adjust the contents of the dictionary through such iterative revisions.
\begin{wraptable}{l}{8.0cm}
%\centering
%\small
\vspace*{-0.6cm}
\resizebox{0.65\textwidth}{!}{
\begin{tabular}{|p{3.2cm}| p{7.2cm}|p{1.5cm}|} \hline
{\bf Entity type} &  {\bf Sample entities} & {\bf \#entities}\\ \hline 
\textbf{Package (PKG)} & \texttt{pypy-configparser}, \texttt{account-plugin-twitter}, \texttt{gtkhtml3.2}, \texttt{pdfcrack}, \texttt{libfields-camlp4-dev-pf4q7} & 140062 \\ \hline
\textbf{Operating System (OS)} & \texttt{SymbOS}, \texttt{Unix System III}, \texttt{NOS}, \texttt{Windows}, \texttt{Cosmic}, \texttt{kubuntu}  & 877 \\ \hline
\textbf{Organization (ORG)} & \texttt{launchpad}, \texttt{bugzilla}, \texttt{sourceforge}, \texttt{nokia}, \texttt{HP} & 379 \\ \hline
\textbf{Command (CMD)} & \texttt{alias}, \texttt{arch}, \texttt{bzip}, \texttt{cat}, \texttt{clear} & 141 \\ \hline
\textbf{Error (ERR)} & \texttt{No such process}, \texttt{No child processes}, \texttt{EFAULT}, \texttt{Bad address}, \texttt{EFBIG} & 124 \\ \hline
\textbf{Extension (EXT)} & \texttt{.asm}, \texttt{.gz}, \texttt{.html}, \texttt{.log}, \texttt{.php} & 76 \\ \hline
\textbf{Peripheral (PRP)} & \texttt{keyboard}, \texttt{mouse}, \texttt{printer}, \texttt{scanner}, \texttt{microphone} & 23 \\ \hline
\textbf{Software component (SOC)} & \texttt{bios}, \texttt{driver}, \texttt{ui}, \texttt{ntfs}, \texttt{fat32} & 12 \\ \hline
\textbf{Architecture (ARC)} & \texttt{x86}, \texttt{x64}, \texttt{32-bit}, \texttt{64-bit}, \texttt{amd64} & 7 \\ \hline 
\end{tabular}
}
\vspace*{-0.3cm}
\caption{The different entity types their count and example entities in our dictionary.}%\footnotesize
\label{tab:entExmpl}
\vspace*{-0.5cm}
\end{wraptable}
\noindent \textbf{Human intervention}: In this step, we avail human intervention to identify cases where a phrase has been incorrectly tagged as an entity. Active learning is particularly useful here, as the model can learn from the feedback provided by the human expert, improving its future predictions (see Appendix~\ref{sec:act-lev-ann}). We randomly sample a few automatically annotated bugs from all the bugs and check which phrases could potentially be non-entities.
We encounter a challenge as the number of entities for categories like software components and peripherals is quite low in our entity list. To expand this dictionary, we utilize the software tool, \textsc{TagMe}\footnote{https://sobigdata.d4science.org/web/tagme/tagme-help}, to extract mentions from Wikipedia and consider them as entities. Along with that process, after receiving the annotated data from Stage 1 and Parts-of-speech filtered bugs (L) also with their IO mentions, a binary RoBERTa-based classifier is trained. We sample 100 bugs on a yearly basis, with our sample spanning from 2004 to 2019. Subsequently, we take up 100*16 = 1600 bugs from TagMe but not included in Stage 1 and feed them into our model for classification as either ``entity” or ``non-entity”. If the confidence for an entity being identified is 50\% or greater, it undergoes manual labeling by a human to categorize it into one of the nine predefined entity types. Once the manual annotation is completed, the newly labeled data (d) is integrated into L, resulting in an updated dataset, (L~$\cup$~d). The model is then retrained using this revised dataset, and the entire procedure is reiterated until the stopping criterion is met. The stopping criterion is defined as the absence of data in the list of entities identified by TagMe.
%~\textcolor{red}{We encounter a challenge as the number of entities for categories like software components and peripherals is quite low in our entity list. To expand this dictionary, we utilize the software tool, \textsc{TagMe}\footnote{https://sobigdata.d4science.org/web/tagme/tagme-help}, to extract mentions from Wikipedia and consider them as entities. This task is carried out in two steps. Considering the human involvement, we sample 100 bugs on a yearly basis, with our sample spanning from 2004 to 2019. These bugs are then processed through \textsc{TagMe} to obtain mentions. To reduce manual effort, we prepare a set of filtered mentions, i.e., those that were not tagged with any entity type during dictionary matching.} 
Finally, our requirement for human intervention end up with around $\sim$3682 filtered mentions, a relatively small number considering the size of our dataset which contains 170K bugs (equivalent to $\sim$1.2 million entities). Distributions of identified entities are shown in Figure~\ref{fig:galaxy}. %\am{The plot should have prop}%\am{I do not understand this figure? Why would you compare \textsc{DistALANER} with \textsc{TagMe}? You are using \textsc{TagMe} to improve \textsc{DistALANER} already?}\SB{Sorry my bad. It is not DistAlNER vs Tagme. It should be the number of entities from Stage I and Extended entities from stage II(Tagme part)}.

\if{0}\noindent\textit{\textbf{RQ: Why LLM can’t be utilize for generating distant labels?}}\\
\noindent\textit{In our current research, we employ a consistent instruction-based prompt to extract and tag entities from a given paragraph. The goal is to return these tagged entities, along with their start and end indices, in a JSON format. The paragraph is analyzed to identify entities that can be classified into one of the following categories: package, operating system, organization, command, error, extension, peripheral, software component, or architecture. We evaluate the outputs from three different language models: GPT-3.5-Turbo, GPT-4, and Google BARD. Through our analysis, we make several observations: (a) GPT-4 fails to generate the complete JSON format, unlike GPT-3.5-Turbo and Google BARD which successfully return the entire format. (b) The output pattern from GPT-4 and Google BARD exhibits more consistency than that of GPT-3.5-Turbo. For instance, when generating entities using GPT-4 and Google BARD, the model returns the entity, its type, and its start and end indices. Conversely, GPT-3.5-Turbo occasionally returns the entity type and indices without the text itself, or merely refers to specific lines in the text, leading to inconsistencies in its output structure. (c) In terms of content, both GPT-3.5-Turbo and GPT-4 extract more accurate entities compared to Google BARD. However, the indices produced by GPT-3.5-Turbo and GPT-4 tend to contain more errors compared to those generated by Google BARD. (d) We notice that all three models sometimes invent their own entity types, which they subsequently return in the output. (e) Finally, it often come with substantial financial considerations, particularly for smaller entities or individual researchers. The costs associated with usage, data storage, and computation can escalate quickly, creating a barrier to access. }\rh{We have to refer to appendix. Also, how to represent the output for three bugs in appendix?}\SB{Creating the big table. May take some time.}\fi

\subsection{Stage 3: The NER model}
\if{0}\am{This part is not novel and we might face a novelty question in the rebuttal. There is nothing software system specific in what follows. It is a mere repetition of the literature.}
In Stage 3, we describe the NER model. We mainly use the SOTA transformer-based CRF models which comprises three steps -- (a) the encoder, (b) the CRF layer, and the (c) log-likelihood loss. The steps are briefly outlined below. \\
\noindent \textbf{Encoder}: In order to encode the texts of the bugs, we use transformer based models and further obtain the embedding for each token in the text.\\
\noindent \textbf{CRF}: Conditional Random Field (CRF) is a probabilistic graphical model commonly used for sequence tagging tasks such as NER. It models the sequence labeling task taking into account the dependencies among the neighboring labels.
$P(Y|X) = \frac{1}{Z} \exp\left(\sum_{i=1}^{n} \theta^\top f(x_i, y_i) + \sum_{i=1}^{n-1} \psi^\top g(y_i, y_{i+1})\right)$
where $P(Y|X)$ is the conditional probability of label sequence $Y$ given input sequence $X$, $Z$ is the normalization factor to ensure the probabilities sum up to 1, $\theta$ and $\psi$ are parameter vectors, $f(x_i, y_i)$ and $g(y_i, y_{i+1})$ are feature functions that capture the compatibility between input-output pairs.\\
\noindent \textbf{Loss function}: We use negative log likelihood loss function in CRF module.
\[
\text{{NLL Loss}} = - \sum_{t=1}^{T} \log\left(\frac{{\exp\left(s_{y_t}(t)\right)}}{{\sum_{c=1}^{C}\exp\left(s_c(t)\right)}}\right)
\]
In this equation, \(T\) represents the sequence length, \(y_t\) denotes the true label of the \(t\)-th position in the sequence, and \(s_c(t)\) is the score of the \(c\)-th label at the \(t\)-th position. The numerator \(\exp\left(s_{y_t}(t)\right)\) corresponds to the score of the true label, and the denominator \(\sum_{c=1}^{C}\exp\left(s_c(t)\right)\) represents the sum of scores over all possible labels at the \(t\)-th position. The logarithm function \(\log(\cdot)\) computes the natural logarithm, and the negative sign \(-\) is applied to obtain the negative log-likelihood loss.\\

\SB{Start of new stage 3}\
\fi
We use a variety of NER models for obtaining the final tags. These include Linear-CRF~\cite{Lafferty:2001}, BilSTM-CRF~\cite{Zhiheng:2015}, BERT-CRF~\cite{souza2020portuguese}, BERT-NER~\cite{liu2021nerbert}, RoBERTa-CRF~\cite{jurkiewicz-etal-2020-applicaai}, SpanBERT-CRF~\cite{portelli2021improving}, and SoftNER~\cite{tabassum-etal-2020-code}. All models are implemented to execute task-specific actions, with their performance evaluated using precision, recall, and F1-score. However, recall is highlighted in this paper as a representational metric for all classes in NER tasks.
%The Linear-CRF model is employed first, utilizing a linear chain of Conditional Random Fields (CRF) to identify the best label sequence for the text. The model is trained using the standard parameters in the CRF library. We then implement the BiLSTM-CRF model, which combines BiLSTM and CRF. The BiLSTM segment generates word embedding features, with the CRF component defining the optimal label sequence. Next, the BERT-CRF and BERT-NER models are used. The BERT-CRF model amalgamates the BERT model with a CRF layer for sequence labeling. In contrast, the BERT-NER model directly uses the BERT model for the NER task. Both models undergo fine-tuning with our collected data. Following this, the Roberta-CRF model is executed, fusing the RoBERTa model with a CRF layer for sequence labeling. The RoBERTa model, an optimized variant of BERT, uses an extensive dataset for training and tweaks the training process for improved results. We also apply the SpanBERT-CRF model that unites the SpanBERT model, a BERT extension emphasizing span-level predictions, and a CRF layer for sequence labeling. This model is trained using the standard settings. For each model implementation, we use a grid search technique to find the optimal hyperparameters (see Appendix~\ref{tab:hyperparameter}). Finally, we employ the SoftNER model to assess its performance with parameter settings as mentioned in the original paper.\\
We train the existing CRF based models using our distantly labelled data obtained from Stage 1 and Stage 2. We note the different hyperparameters for the above models in Appendix.
%All CRF-based models consist of three parts: (a) the encoder, (b) the CRF Layer, and (c) Log-likelihood loss. The encoder first encodes the task, after which the CRF is used for sequence tagging, taking into account the dependencies among neighboring labels. The conditional probability of label sequence $Y$ given input sequence $X$ is represented as $P(Y|X) = \frac{1}{Z} \exp\left(\sum_{i=1}^{n} \theta^\top f(x_i, y_i) + \sum_{i=1}^{n-1} \psi^\top g(y_i, y_{i+1})\right)$, where $Z$ is the normalization factor, $\theta$ and $\psi$ are parameter vectors, and $f(x_i, y_i)$ and $g(y_i, y_{i+1})$ are feature functions. We use the negative log-likelihood loss function in the CRF module, defined as follows: 
% \[
% \text{{NLL Loss}} = - \sum_{t=1}^{T} \log\left(\frac{{\exp\left(s_{y_t}(t)\right)}}{{\sum_{c=1}^{C}\exp\left(s_c(t)\right)}}\right)
% \]
%\[
%\text{NLL}(X_i, Y_i) = -\log \left(\frac{{\exp(\text{Score}(X, Y))}}{{\sum_{Y'} \exp(\text{Score}(X, Y'))}}\right)
%\]
We also use LLMs in a zero-shot setting with instruction based prompt to extract the entities along with its start and end index from the paragraph (see section~\ref{sec:appendix4} for more details)\footnote{We try multiple prompt variants and retain the one that produces the best results.}.
%\SB{End of new stage 3}
%\SB{Possibly if we write this then we can skip the baseline part and incorporate the progressive learning part in this main paper.}
\vspace*{-0.3cm}
\section{Heuristics}\label{sec:appendix1}
%\SB{Added this whole section from Appendix}
We use Part-of-Speech (POS) tagging as a heuristic to discard certain entities. In our process, we filter out the entities based on their POS tags. For example, we discard entities that are typically labeled as conjunctions, interjections, or prepositions as these are less likely to represent valid entities. This strategy ensures that only the most relevant words, such as nouns or proper nouns, are considered for entity recognition. This way, we can reduce noise in the data and improve the performance of our NER model. Furthermore, these POS tag-based heuristics help us in refining our entity list, leading to a more accurate and efficient distant supervision process in NER (See Table~\ref{tab:heuristics} for more samples).
\vspace*{-0.3cm}
\section{Experimental setup}
% \subsection{Training and test data preparation}

\noindent \textbf{Training and test data preparation}: In order to train the NER model in Stage 3, we need training data. For this purpose we select all the bug descriptions from the years 2004 through 2013 inspired by~\cite{10.1007/978-3-030-72113-8_15}. These descriptions, nearly 65K in number are automatically annotated using the first two stages of our framework. Thus, the training data only has auto-curated (aka silver) entity labels without any human involvement. For the test set we consider the data from the years 2016 to 2019. To assess the performance of the model, which has been trained using distantly labeled data, we perform human annotation of a subset of the test set.
\textit{Human annotation}: We present 500 bug descriptions to four domain experts, each claiming over three years of experience in the opensource ecosystem and package management. The resulting inter-annotator agreement among these annotators is 0.625 (see Appendix for more information). For the Launchpad QA dataset, we employ the same group of experts to annotate the entities in 500 question-answer pairs, ensuring consistency in our data annotation process across both datasets.
% \subsection{Baselines}
\noindent \textbf{Baselines}: In addition to the NER models and zero-shot LLMs, we also use the combined output of Stage 1 and 2 as a baseline.\\
\if{0}\subsection{Parameter settings}
\SB{Possibly will go to appendix}\am{yes}
\begin{table}[!h]
%\tiny
\scalebox{0.41}{
\begin{tabular}{|l|l|}
\hline
\multicolumn{1}{|c|}{\textbf{Methods}} & \multicolumn{1}{c|}{\textbf{Hyperparameters}}                                                                                                                                                                                     \\ \hline
Linear CRF                             & "Passive Aggressive" algorithm, 150 iterations                                                                                                                                                                                         \\ \hline
BiLSTM CRF                             & \begin{tabular}[c]{@{}l@{}}embedding dim= 768, BiLSTM dim = 256, LEARNING\_WEIGHT = 5e-2 \\ WEIGHT\_DECAY = 1e-4, epochs = 3\end{tabular}                                                                                              \\ \hline
Bert NER                               & \begin{tabular}[c]{@{}l@{}}dropout = 0.1, max\_seq = 512, AdamW, epochs = 15, lr: 5.0e-06, batch\_size = 32\\ lr\_scheduler:\\     end\_factor: 0.0,start\_factor: 1.0,total\_iters: 25, type: LinearLR\end{tabular}  \\ \hline
Bert CRF                               & \begin{tabular}[c]{@{}l@{}}dropout = 0.1, max\_seq = 512, AdamW, epochs = 15, lr: 5.0e-06, batch\_size = 32\\ lr\_scheduler:\\     end\_factor: 0.0,start\_factor: 1.0,total\_iters: 25,type: LinearLR\end{tabular}  \\ \hline
Partial Bert CRF                       & \begin{tabular}[c]{@{}l@{}}dropout = 0.1, max\_seq = 512, AdamW, epochs = 15, lr: 5.0e-06, batch\_size = 32\\ lr\_scheduler:\\     end\_factor: 0.0,start\_factor: 1.0,total\_iters: 25,type: LinearLR\end{tabular}  \\ \hline
spanBert CRF                           & \begin{tabular}[c]{@{}l@{}}dropout = 0.1, max\_seq = 512, AdamW, epochs = 5, lr: 5.0e-06, batch\_size = 11, \\ lr\_scheduler:\\     end\_factor: 0.0,start\_factor: 1.0,total\_iters: 30,type: LinearLR\end{tabular} \\ \hline
SoftNER                                & epochs = 10, bert-base-uncased, max\_seq = 512, lr = 5e-5,epsilon for adam optimiser = 1e-8                                                                                                                                            \\ \hline
Roberta CRF                            & \begin{tabular}[c]{@{}l@{}}dropout = 0.1, max\_seq = 512, AdamW, epochs = 15, lr: 5.0e-06, batch\_size = 32\\ lr\_scheduler:\\     end\_factor: 0.0,start\_factor: 1.0,total\_iters: 25,type: LinearLR\end{tabular}  \\ \hline
Pretrained Roberta CRF                 & \begin{tabular}[c]{@{}l@{}}dropout = 0.1, max\_seq = 512, AdamW, epochs = 5, lr: 5.0e-06, batch\_size = 32\\ lr\_scheduler:\\     end\_factor: 0.0,start\_factor: 1.0,total\_iters: 25,type: LinearLR\end{tabular}   \\ \hline
\end{tabular}
}
\caption{Hyperparameters}
\label{tab:hyperparameter}
\end{table}\fi
%\subsection{Metric for evaluation}
%\vspace*{-0.4cm}
\noindent\textbf{Metric for evaluation}: To evaluate all the models, we compute the recall rate inspired by~\cite{tu-lignos-2021-tmr}. Recall rate is the proportion of actual entity types in ground truth that are accurately predicted. We use this metric rather than the F1 score for classwise evaluation in the main table in cognizance of the fact that human annotations are sometimes incomplete while the model is able to generate the correct entity type (see Section~\ref{sec:results}). Precisely, when dealing with software-related texts, overlooking an entity can lead to the omission of vital data. For instance, not recognizing an error code, a software package, or a particular function name can change the outcome from comprehending and addressing a problem to failing to do so. This makes capturing as many relevant entities as we can, which recall assesses, crucial. Further, software-related documents typically present crucial details that require exhaustive extraction for subsequent applications like bug tracking, requirement analysis, or code generation. Overlooking an entity in these contexts can have an adverse impact, underscoring the importance of recall as a key metric. Nevertheless, in order to make the results complete we also subsequently measure and report the recall and F1 score.
\section{Results}\label{sec:results}
\begin{wraptable}{r}{8.0cm}
\vspace*{-0.8cm}
\centering
\small
\scalebox{0.55}{
\begin{tabular}{|c|c|c|}
\hline
\cellcolor[HTML]{FFFFFF}\textbf{Sample Bugs} &
  \textbf{Sample Tag Heuristics} &
  \textbf{Conversion} \\ \hline
"Some selected error messages from the time of session login" &
  \begin{tabular}[c]{@{}c@{}}Messages - (NN, NNS, IN)\\ Time - (DT, NN, IN)\end{tabular} &
  \begin{tabular}[c]{@{}c@{}}Error -\textgreater O\\ Package -\textgreater O\end{tabular} \\ \hline
\begin{tabular}[c]{@{}c@{}}"This results in a serious compromise on the possibility of \\ running remote displays systems on ubuntu. In fact, the latter can only \\ rely on Xvfb with a less than optimal experience."\end{tabular} &
  Less - (DT, JJR, IN) &
  Package -\textgreater O \\ \hline
\begin{tabular}[c]{@{}c@{}}"SST will fail if donor has to send keyring. \\ Looks like the donor is trying to send the file \\ while so cat is still opening port 4444 on joiner..."\end{tabular} &
  \begin{tabular}[c]{@{}c@{}}File - (DT, NN, IN)\\ File - (NNP, NN, IN)\\ File - (JJ, NN, IN)\end{tabular} &
  \begin{tabular}[c]{@{}c@{}}Package -\textgreater O\\ Package -\textgreater O\\ Package -\textgreater O\end{tabular} \\ \hline
\end{tabular}
}
\caption{\footnotesize Samples of some heuristics to discard wrongly identified entities.}
\label{tab:heuristics}
\vspace*{-0.6cm}
%\end{table*}
\end{wraptable}
We evaluate all the baselines for Ubuntu bug dataset, Launchpad QA, Fedora forum and Linux community dataset. We show classwise recall values for Ubuntu bug dataset and Launchpad QA in Table~\ref{tab:result} while the overall performance in terms of macro-F1 score is shown in Table~\ref{tab:my_overall_table}. Our experiments involve two different setups -- (a) Human-Induced Training (HIn), and (b) Human-Only Training (HOn). %\am{what do you mean by human induced and human only training? Also describe your metric of evaluation in 1-2 lines.}
\begin{table*}[t]
%\centering
%\small
\resizebox{1.0\textwidth}{!}{
\begin{tabular}{|c|c|c|c|c|c|c|c|c|c|c|c|c|c|c|c|c|c|c|c|c|} \hline%{|p{3.5cm}| p{2cm}| p{3 cm}|}
\multirow{2}{*}{\bf Methods} & \multicolumn{2}{c|}{\bf ARC} & \multicolumn{2}{c|}{\bf CMD} & \multicolumn{2}{c|}{\bf ERR} & \multicolumn{2}{c|}{\bf EXT} & \multicolumn{2}{c|}{\bf OS} & \multicolumn{2}{c|}{\bf ORG} & \multicolumn{2}{c|}{\bf PKG} & \multicolumn{2}{c|}{\bf PRP} & \multicolumn{2}{c|}{\bf SOC} & \multicolumn{2}{c|}{\bf Overall}\\ \cline{2-21}
 & HIn & HOn & HIn & HOn & HIn & HOn & HIn & HOn & HIn & HOn & HIn & HOn & HIn & HOn & HIn & HOn & HIn & HOn & HIn & HOn \\ \cline{1-21} 
 \multirow{2}{*}{Direct matching} & 0.966 & --- & \cellcolor{green!25}0.053 & --- & \cellcolor{green!25}0.154 & --- & 0.111 & --- & 0.446 & --- & \cellcolor{green!25}0.541 & --- & \cellcolor{green!25}0.774 & ---- & 0.148 & --- & 0.110 & --- & 0.400 & ---\\ \cdashline{2-21}%\cline{2-11}
 & --- & --- & --- & --- & --- & --- & --- & --- & --- & --- & --- & --- & --- & --- & --- & --- & --- & --- & --- & ---  \\ \cline{1-21} 
  \multirow{2}{*}{GPT-3.5-Turbo} & 0.002 & ---& 0.002 &---& 0.001 &---& 0 &---& 0.004 &---& 0 &---& 0 &---& 0.005 &---& 0 &---& 0.002 & --- \\ \cdashline{2-21}%\cline{2-11}
 & --- & --- & --- & --- & --- & --- & --- & --- & --- & --- & --- & --- & --- & --- & --- & --- & --- & --- & --- & --- \\ \cline{1-21} 
  \multirow{2}{*}{GPT-4} & 0.032 & ---& \cellcolor{green!8}0.052 &---& 0.0111 &---& 0.311 &---& 0.104 &---& 0.202 &---& 0.003 &---& 0.121 &---& 0.023 &---& 0.092 & ---\\ \cdashline{2-21}%\cline{2-11}
   & --- & --- & --- & --- & --- & --- & --- & --- & --- & --- & --- & --- & --- & --- & --- & --- & --- & --- & --- & --- \\ \cline{1-21}  
  \multirow{2}{*}{Google BARD} & 0.002 & ---& 0.012 &---& 0 &---& 0 &---& 0.002 &---& 0 &---& 0 &---& 0.002 &---& 0 &---& 0.001 & ---\\ \cdashline{2-21}%\cline{2-11}
   & --- & --- & --- & --- & --- & --- & --- & --- & --- & --- & --- & --- & --- & --- & --- & --- & --- & --- & --- & ---  \\ \cline{1-21}
   \multirow{2}{*}{UniversalNER} & 0.133 & ---& \cellcolor{green!25}0.053 &---& 0.034 &---& 0.532 &---& 0.101 &---& 0.367 &---& 0.036 &---& 0.144 &---& 0.081 &---& 0.168 & ---\\ \cdashline{2-21}%\cline{2-11}
   & --- & --- & --- & --- & --- & --- & --- & --- & --- & --- & --- & --- & --- & --- & --- & --- & --- & --- & --- & ---  \\ \cline{1-21} 
 \multirow{2}{*}{Linear-CRF} & 0.960 &  \cellcolor{green!8}0.624 & \cellcolor{green!25}0.053 & \cellcolor{green!8}0.155 & \cellcolor{green!8}0.151 & 0.030 & 0.630 & \cellcolor{green!8}0.074 & 0.676 & \cellcolor{green!8}0.680 & \cellcolor{green!8}0.538 & \cellcolor{green!25}0.337 & 0.737 & \cellcolor{green!25}0.347& 0.147 & 0.0317 & 0.120 & \cellcolor{green!8}0.129 & 0.443 & \cellcolor{green!25}0.298\\ \cdashline{2-21}
& \cellcolor{blue!25}0.756 & \cellcolor{blue!25}0.292 & 0.109 & 0.073 & \cellcolor{blue!25}0.418 & 0.012 & 0.264 & 0.022 & \cellcolor{blue!8}0.414 & 0.087 & 0.510 & \cellcolor{blue!25}0.227 & 0.465 & 0.014 & 0.278 & 0.002 & \cellcolor{blue!25}0.362 & \cellcolor{blue!8}0.028 & \cellcolor{blue!8}0.375 & 0.062  \\ \cline{1-21}
 \multirow{2}{*}{BiLSTM-CRF} &  0.698 & \cellcolor{green!25}0.643 & 0.06 &0.026& 0.016 &0.001& 0.358 &0 & 0.654 &\cellcolor{green!25}0.768& 0.347 &0.085& 0.450 &\cellcolor{green!8}0.304&  0.121 &0.081& 0.004 &0.019& 0.314 &  \cellcolor{green!8}0.262\\ \cdashline{2-21}
 & 0.230 & \cellcolor{blue!8}0.141 & 0.053 & 0.017 & 0 & 0 & 0.080 & 0 & 0.103 & 0.145 & 0.44 & 0.110 & 0.113 & \cellcolor{blue!25}0.053  & 0.105 & 0.013 & 0.034 & 0.001 & 0.111 &  0.053\\ \cline{1-21}
 \multirow{2}{*}{BERT-NER} & \cellcolor{green!8}0.968 & 0.026 & 0.051 & 0.036 & 0.142 & \cellcolor{green!8}0.077 & \cellcolor{green!8}0.833 & 0.047 & \cellcolor{green!8}0.729 & 0.261 & 0.500 & \cellcolor{green!8}0.203 & 0.732 & 0.044 & 0.150 & \cellcolor{green!25}0.183 & \cellcolor{green!8}0.124 & \cellcolor{green!25}0.135 & \cellcolor{green!8}0.461 & 0.126\\ \cdashline{2-21}
 & 0.560 & 0.037 & 0.127 & 0.034 & 0.328 & \cellcolor{blue!8}0.042 & 0.492 & \cellcolor{blue!8}0.063 & 0.405 & \cellcolor{blue!8}0.165 & 0.583 & \cellcolor{blue!8}0.167 & \cellcolor{blue!25}0.664 & \cellcolor{blue!8}0.033 & \cellcolor{blue!25}0.306 & \cellcolor{blue!25}0.091 & 0.262 & \cellcolor{blue!25}0.128 & 0.378 & 0.094 \\ \cline{1-21} 
\multirow{2}{*}{BERT-CRF*} & \cellcolor{green!25}0.970 & 0.002 & \cellcolor{green!25}0.053 & 0.033 & 0.144 & 0.076 & \cellcolor{green!8}0.833 & 0.047 & \cellcolor{green!25}0.792 & 0.208 & 0.500 & 0.191 & 0.766 & 0.085 & \cellcolor{green!8}0.153 & \cellcolor{green!8}0.113 & \cellcolor{green!8}0.124 & 0.123 & \cellcolor{green!25}0.481 & 0.106\\ \cdashline{2-21}
 & 0.570 & 0 & \cellcolor{blue!25}0.129 & \cellcolor{blue!8}0.059 & 0.328 & 0.038 & \cellcolor{blue!8}0.539 & 0.031 & \cellcolor{blue!25}0.428 & \cellcolor{blue!25}0.194 & 0.590 & 0.100 & 0.65 & \cellcolor{blue!8}0.033 & \cellcolor{blue!8}0.304 & \cellcolor{blue!8}0.083 & 0.262 & \cellcolor{blue!25}0.128 & \cellcolor{blue!25}0.381 & \cellcolor{blue!8}0.097 \\ \cline{1-21}
 \multirow{2}{*}{RoBERTa-CRF} & 0.923 & 0 & 0.047 & \cellcolor{green!25}0.681 & 0.141 & \cellcolor{green!25}0.151 & 0.778 & \cellcolor{green!25}0.266 & \cellcolor{green!8}0.729 & 0.001 & 0.489 & 0.033 & 0.710 & 0.002 & 0.150 & 0.014 & 0.116 & 0 & 0.452 & 0.112\\ \cdashline{2-21}
 & \cellcolor{blue!8}0.580 & 0 & \cellcolor{blue!8}0.128 & \cellcolor{blue!25}0.608 & 0.323 & \cellcolor{blue!25}0.281 & 0.460 & \cellcolor{blue!25}0.460 & 0.390 & 0  & \cellcolor{blue!8}0.568 & 0.026 & 0.588 & 0.003 & 0.299 & 0.041 & 0.258 & 0.001 & 0.361 & \cellcolor{blue!25}0.109 \\ \cline{1-21} 
% \multirow{2}{*}{Roberta-CRF-PT} & && && && && && &&  && && && &\\ \cdashline{2-21}
%  &  & && &&  &&  &&  && &&  &&  &&  &&  \\ \cline{1-21} 
\multirow{2}{*}{SpanBERT-CRF} & 0.966 & 0 & 0.043 & 0 & 0.139 & 0 & \cellcolor{green!25}0.846 & 0 & 0.445 & 0 & 0.475 & 0 & \cellcolor{green!8}0.771 & 0 & \cellcolor{green!25}0.155 & 0 & \cellcolor{green!25}0.129 & 0 & 0.396 & 0 \\ \cdashline{2-21}
 & 0.566 & 0 & 0.123 & 0& \cellcolor{blue!8}0.330 & 0 & 0.532 & 0 & 0.412 & 0 & \cellcolor{blue!25}0.597 & 0 & \cellcolor{blue!8}0.600 & 0 & 0.295 & 0 & \cellcolor{blue!8}0.264 & 0 & 0.367 & 0 \\ \cline{1-21} 
\multirow{2}{*}{SoftNER} & 0.805 & 0& 0.045 &0 & 0.126 &0 & 0.680 &0 & 0.711 &0 & 0.494 &0 & 0.598 &0& 0.120 & 0& 0.112 & 0& 0.416 &0\\ \cdashline{2-21}
 & 0.568 & 0 & 0.118 & 0 & 0.294 & 0 & \cellcolor{blue!25}0.568 & 0 & 0.391 & 0 & 0.520 & 0 & 0.529 & 0 & 0.294 & 0 & 0.256 & 0 & 0.346 & 0 \\ \cline{1-21} 
\end{tabular}
}  
\caption{\footnotesize Recall rate for named entity recognition approaches. The first sub-row of each row shows the results for the Ubuntu bug dataset while the second sub-row shows the results for the Launchpad QA dataset (inference phase only). The best and the second best results in the first sub-row (bug dataset) are highlighted in \colorbox{green!25}{dark} and \colorbox{green!8}{light} green respectively. The best and the second best results in the second sub-row (QA dataset) are highlighted in \colorbox{blue!25}{dark} and \colorbox{blue!8}{light} purple respectively. BERT-CRF is significantly different from domain specific BERT-NER(except launchpad) and RoBERTa-CRF (*$p < 0.05$). Table~\ref{tab:my_overall_table} compares methods for HIn based on other metrics.}
\label{tab:result}
\vspace*{-0.9cm}
\end{table*}
In the case of a Human-Induced (HIn) configuration, we train the models using all the auto-annotated bug descriptions plus 10\% (or $\sim$50 instances) human-annotated bug descriptions. These examples of human annotated data are incorporated into the training to provide the models with additional knowledge of gold annotations. For Human-Only training (HOn), only 10\% of human annotations are used to train the models (see subsection~\ref{sec:splitHon}). We do not include any auto-annotated data during HOn configuration training. In both setups, 10\% of the human-annotated data is used for training, 20\% for validation, and 70\% for testing.\\
Overall, we observe that the HIn setup outperforms the HOn setup for the NER models for almost all the entity types establishing the effectiveness of the distantly supervised auto-annotations. In the HIn setup, we find that BERT-CRF outperforms other models in overall performance while BERT-NER is the second best. For the entity types \textbf{ARC} and \textbf{PKG}, all models exhibit good performance, with the exception of BiLSTM-CRF. However, when it comes to identifying \textbf{CMD}, \textbf{ERR}, and \textbf{SOC}, all models face challenges, with BiLSTM-CRF reporting the worst performance. We observe \textbf{CMD}, \textbf{ERR} have relatively lengthy entity names (three to four words) compared to the other entity types; consequently, in the majority of cases, NER models fail to identify the correct entity types for these large names. The most consistent results across all entity types primarily come from BERT-CRF and BERT-NER. In the HOn setup, we note that overall recall is high for Linear-CRF and BiLSTM-CRF compared to other models. However, when we look at entity-wise recall, Linear-CRF, BERT-CRF, and BERT-NER perform better than the other models. Interestingly, SoftNER and SpanBERT-CRF struggle to learn in this setup. For both the setups, we use dark green to denote the best-performing value and light green for the second position across all the models (see Table~\ref{tab:result}).\\
Further, we examine the trained models' performance on the Launchpad dataset. Here, we use the trained models (trained using bug descriptions in both HIn and HOn setups) only for inference purposes, i.e., we do not train the models on the Launchpad dataset. The key idea is to test the performance in a zero-shot transfer setup. Here, BERT-CRF exhibits the best overall performance while BERT-NER and Linear-CRF are in the second best position. In Table~\ref{tab:result}, we report the best model performance in dark purple while light purple denotes the second best. A detailed analysis of the error cases are presented in section~\ref{sec:erroranalysis}.\\
% \usepackage{multirow}
\begin{table*}[t]
%\vspace*{-0.7cm}
\resizebox{1.0\textwidth}{!}{
\begin{tabular}{|c|ccccccc||cccc|}
\hline
\multirow{2}{*}{\textbf{Methods}} &
  \multicolumn{7}{c||}{\textbf{Pre-LLM era}} &
  \multicolumn{4}{c|}{\textbf{LLM era}} \\ \cline{2-12} 
 &
  \multicolumn{1}{c|}{\textbf{Linear-CRF}} &
  \multicolumn{1}{c|}{\textbf{BiLSTM-CRF}} &
  \multicolumn{1}{c|}{\textbf{BERT-NER}} &
  \multicolumn{1}{c|}{\textbf{BERT-CRF}} &
  \multicolumn{1}{c|}{\textbf{RoBERTa-CRF}} &
  \multicolumn{1}{c|}{\textbf{spanBERT-CRF}} &
  \textbf{SoftNER} &
  \multicolumn{1}{c|}{\textbf{GPT-3.5-Turbo}} &
  \multicolumn{1}{c|}{\textbf{GPT-4}} &
  \multicolumn{1}{c|}{\textbf{Google BARD}} &
  \textbf{UniversalNER} \\ \hline
\textbf{\begin{tabular}[c]{@{}c@{}}Ubuntu\\ (Bug)\end{tabular}} &
  \multicolumn{1}{c|}{0.410} &
  \multicolumn{1}{c|}{0.290} &
  \multicolumn{1}{c|}{0.424} &
  \multicolumn{1}{c|}{\cellcolor{blue!25}0.471} &
  \multicolumn{1}{c|}{\cellcolor{blue!8}0.448} &
  \multicolumn{1}{c|}{0.350} &
  0.396 &
  \multicolumn{1}{c|}{0.002} &
  \multicolumn{1}{c|}{0.091} &
  \multicolumn{1}{c|}{0.001} &
  0.149 \\ \hline
\textbf{\begin{tabular}[c]{@{}c@{}}Launchpad\\ (QA)\end{tabular}} &
  \multicolumn{1}{c|}{0.354} &
  \multicolumn{1}{c|}{0.090} &
  \multicolumn{1}{c|}{\cellcolor{blue!25}0.366} &
  \multicolumn{1}{c|}{\cellcolor{blue!8}0.360} &
  \multicolumn{1}{c|}{0.342} &
  \multicolumn{1}{c|}{0.330} &
  0.319 &
  \multicolumn{1}{c|}{0.001} &
  \multicolumn{1}{c|}{0.082} &
  \multicolumn{1}{c|}{0.000} &
  0.193 \\ \hline
\textbf{\begin{tabular}[c]{@{}c@{}}Fedora\\ (CQA)\end{tabular}} &
  \multicolumn{1}{c|}{0.403} &
  \multicolumn{1}{c|}{0.323} &
  \multicolumn{1}{c|}{\cellcolor{blue!8}0.429} &
  \multicolumn{1}{c|}{\cellcolor{blue!25}0.495} &
  \multicolumn{1}{c|}{0.417} &
  \multicolumn{1}{c|}{0.213} &
  0.314 &
  \multicolumn{1}{c|}{0.009} &
  \multicolumn{1}{c|}{0.018} &
  \multicolumn{1}{c|}{0.003} &
  0.191 \\ \hline
\textbf{\begin{tabular}[c]{@{}c@{}}Linux\\ (CQA)\end{tabular}} &
  \multicolumn{1}{c|}{0.441} &
  \multicolumn{1}{c|}{0.285} &
  \multicolumn{1}{c|}{0.449} &
  \multicolumn{1}{c|}{\cellcolor{blue!25}0.507} &
  \multicolumn{1}{c|}{\cellcolor{blue!8}0.477} &
  \multicolumn{1}{c|}{0.302} &
  0.371 &
  \multicolumn{1}{c|}{0.009} &
  \multicolumn{1}{c|}{0.033} &
  \multicolumn{1}{c|}{0.003} &
  0.204 \\ \hline
\end{tabular}
}
\caption{\footnotesize Comparison of methods for HIn based on macro-F1 Scores. BERT-CRF is significantly different from domain specific BERT-NER(except launchpad) and RoBERTa-CRF (*$p < 0.05$). The best and the second best results are highlighted in \colorbox{blue!25}{dark} and \colorbox{blue!8}{light} purple respectively.}
\label{tab:my_overall_table}
\vspace*{-0.7cm}
\end{table*}
For evaluating overall results in terms of macro-F1 score we divide methods in two eras: the pre-LLM era and the LLM era. In the pre-LLM era, BERT-CRF consistently demonstrates superior performance across different datasets (Ubuntu (Bug), Launchpad (QA), Fedora (CQA), and Linux (CQA)), securing the highest macro-F1 scores highlighted in dark purple. Following closely, RoBERTa-CRF emerges as the second-best method in terms of performance, indicated by light purple highlights in the Ubuntu (Bug) and Linux (CQA) datasets. Transitioning to the LLM era, we notice a stark contrast in performance. Notably, the scores drastically drop, with GPT-3.5-Turbo, GPT-4, and Google BARD exhibiting significantly lower macro-F1 scores across all datasets, suggesting that despite their advanced capabilities, these models may not be directly optimized for the specific task of NER as compared to their predecessors in the pre-LLM era. However, UniversalNER demonstrates relatively better performance in this era, albeit still not reaching the effectiveness of the pre-LLM methods.
\vspace*{-0.25cm}
\subsection{Progressive learning}
\begin{wrapfigure}{r}{0.45\textwidth} %this figure will be at the right
    \centering
    \vspace*{-0.75cm}
    \includegraphics[width=5cm]{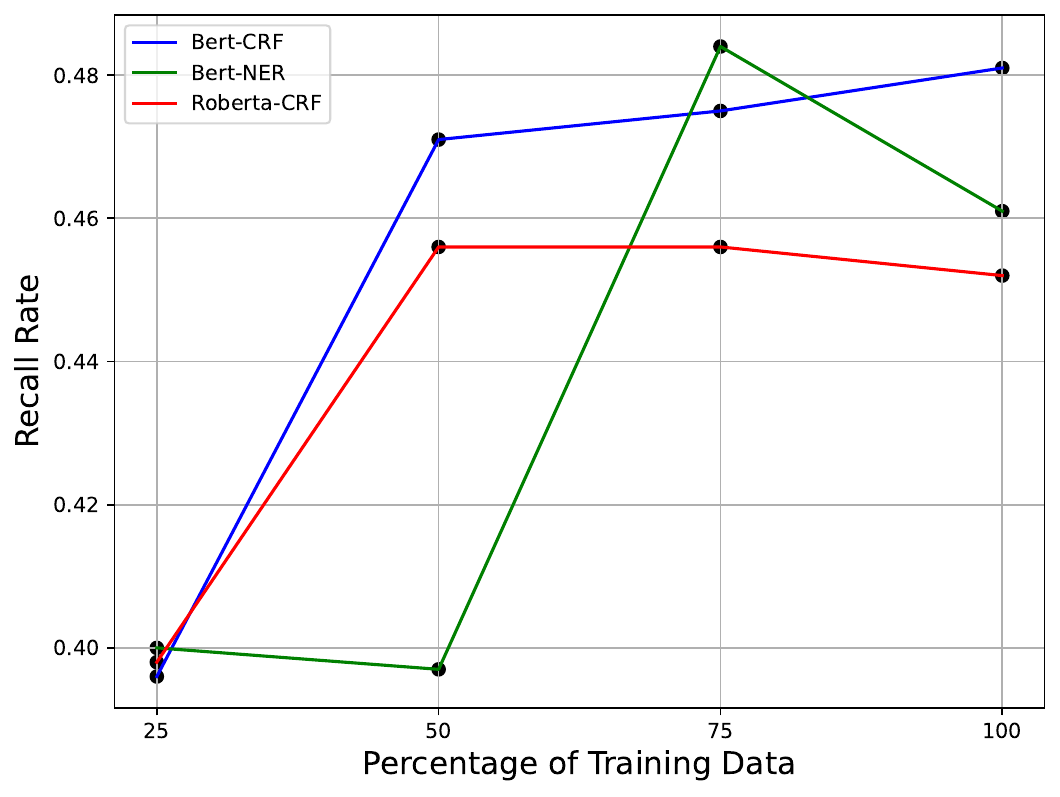}
    \caption{\footnotesize Recall rate for varying percentage of training data.}
    \label{fig:percentageDist}
\vspace*{-0.7cm}
\end{wrapfigure}

Progressive learning~\cite{chatterjee2017progressive} involves incrementally training machine learning models with increasing amounts of data. In our experiment, we divide the data into 25\%, 50\%, and 75\% segments among equally distributed entities. Figure~\ref{fig:percentageDist} illustrates the Recall rate distribution across different percentages of training data. We focus on the top three performing models and observe that BERT-CRF consistently improves with additional data, ultimately achieving the best performance when trained with 100\% of the data.
% \begin{table*}[!h]
% \centering
% \small
% \begin{tabular}{|c|c|c|c|c|c|c|c|c|c|c|} \hline%{|p{3.5cm}| p{2cm}| p{3 cm}|}
% {\bf Methods} & {\bf ARC} & {\bf CMD} & {\bf ERR} & {\bf EXT} & {\bf OS} & {\bf ORG} & {\bf PKG} & {\bf PRP} & {\bf SOC} & {\bf Overall}\\ \hline 
% Direct matching  &&&&&&&  &&&\\ \hline
% Linear-CRF &&&&&&&  &&&\\ \hline
% BiLSTM-CRF &&&&&&&  &&&\\ \hline
% BERT-NER &&&&&&&  &&&\\ \hline
% BERT-CRF &&&&&&&  &&& \\ \hline
% Roberta-CRF &&&&&&&  &&&\\ \hline
% Roberta-CRF-PT &&&&&&&  &&& \\ \hline
% SpanBERT-CRF &&&&&&&  &&&\\ \hline 
% SoftNER &&&&&&&  &&&\\ \hline
% \end{tabular}
% \caption{\label{tab:dataset} Recall rate for named entity recognition models for {\bf launchpad question answering dataset}}%\footnotesize
% \end{table*}
%\vspace*{-0.9cm}
\subsection{Motivation of different split of HOn}\label{sec:splitHon}
We attempt to determine how the model performs using various data splits, primarily to verify consistent result trends. For a more insightful comparison, we divide the data into a randomly 50:10:40 (Train: Valid: Test) ratio. Table~\ref{tab:performance_metrics} presents the performance based on this split. Clearly, as Table~\ref{tab:performance_metrics} indicates, our findings align with the trends identified in the main paper for BERT-CRF. We compare BERT-CRF and Linear-CRF as Linear-CRF ranks as either top or second-best competitor considering all entities, making it one of the prime competitor to BERT-CRF.

% \begin{table*}[h!]
%     \centering
%     \small
%     \scalebox{0.95}{
%     \begin{tabular}{|c|c|c|c|c|c|c|c|c|c|c|}
%     \hline
%     \textbf{Method} & \textbf{ARC} & \textbf{CMD} & \textbf{ERR} & \textbf{EXT} & \textbf{OS} & \textbf{ORG} & \textbf{PKG} & \textbf{PRP} & \textbf{SOC} & \textbf{Overall} \\
%     \hline
%     BERT CRF (HOn) (Bug) & 0.040 & 0.044 & 0.089 & 0 & 0.319 & 0.183 & 0.024 & 0.179 & 0.138 & 0.137 \\
%     \hline
%     BERT CRF (HOn) (Launchpad) & 0.037 & 0.043 & 0.042 & 0.063 & 0.194 & 0.174 & 0.032 & 0.090 & 0.134 & 0.089 \\
%     \hline
%     Linear CRF (HOn) (Bug) & 0.792 & 0.190 & 0.007 & 0.513 & 0.662 & 0.371 & 0.419 & 0.193 & 0.148 & 0.356 \\
%     \hline
%     Linear CRF (HOn) (Launchpad) & 0.088 & 0.140 & 0.028 & 0.218 & 0.181 & 0.027 & 0.029 & 0.009 & 0.019 & 0.068 \\
%     \hline
%     \end{tabular}
%         }
%     \caption{\footnotesize Comparison of performance metrics for BERT-CRF and Linear-CRF methods across various categories using different datasets (Bug and Launchpad).}

%     \label{tab:performance_metrics}
% \end{table*}

% \usepackage{multirow}

\section{Issues in finding distant labels using LLMs}\label{sec:appendix4}
Our research employs a uniform instruction-based approach to extract and tag entities from text, aiming to produce tagged entities with their indices in JSON format, as demonstrated in the Appendix with sample prompts. We assessed outputs from GPT-3.5, GPT-4, Google BARD, and UniversalNER, noting several key findings as follows. \textit{\textbf{(a) Consistency and format}}: GPT-4 and Google BARD show consistent output patterns, unlike GPT-3.5-Turbo, which sometimes misses text or references lines inconsistently. GPT-3.5 and GPT-4 excel in entity accuracy but fall behind in index precision compared to BARD. \textit{\textbf{(b) Invented entities}}: All models occasionally create new, untrained entity types in their outputs. \textit{\textbf{(c) UniversalNER performance}}: Despite being specifically trained, UniversalNER struggles with entity accuracy. \textit{\textbf{(d) Cost efficiency}}: Our distantly supervised approach offers a more cost-effective solution for entity tagging compared to direct model predictions. Some examples in appendix further illustrate these points.

\section{Additional experiments for task-based evaluation}
Relation extraction and NER are closely intertwined in many NLP tasks. They are often referred to as ``sister" problems in the literature. Many studies, including~\cite{wang-etal-2022-named,zhong-chen-2021-frustratingly}, explore these two tasks jointly, highlighting their interdependence. While relation extraction identifies and classifies the relationships between recognized entities, it often serves as a supplementary evaluation, building upon the foundational results of NER models.
\subsection{Relation extraction}
\begin{wraptable}{l}{8.5cm}
\vspace{-0.7cm}
\centering
%\small
\resizebox{.70\textwidth}{!}{
\begin{tabular}{|c|c|c|c|c|c|} \hline
{\bf Method} &  {\bf Overall}& {\it dependency} & {\it affected versions} & {\it cause and effect} & {\it interaction/control}\\ \hline 
BERT-CRF & \cellcolor{cyan!25}0.46 & \cellcolor{cyan!25}0.59 & \cellcolor{cyan!25}0.15 & \cellcolor{cyan!25}0.49 & 0.45 \\ \hline
BERT-NER & \cellcolor{cyan!8}0.44 & \cellcolor{cyan!25}0.59 & 0.12 & \cellcolor{cyan!8}0.45 & \cellcolor{cyan!8}0.46\\ \hline
RoBERTa-CRF & \cellcolor{cyan!8}0.44 & 0.54 & \cellcolor{cyan!8}0.14 & \cellcolor{cyan!25}0.49 & \cellcolor{cyan!8}0.46\\ \hline
Vanilla BERT & 0.42& 0.51 & 0.06 & \cellcolor{cyan!8}0.45 & \cellcolor{cyan!25}0.48\\ \hline
\end{tabular}
}
\caption{\footnotesize Relation extraction performance (F1 scores).}%\footnotesize
\label{tab:relEx} 
\vspace{-0.5cm}
\end{wraptable}
Relation extraction (RE) is the most natural follow-up task of NER. In this case study we attempt to identify the effectiveness of NER in solving the RE task in the context of open source software systems. We identify five broad types of relationships: (a) {\em dependency} -- a dependency relation between two entity types indicates that one entity depends on or relies on the other entity for its proper functioning or execution (e.g. triplets include (\texttt{sane-utils}, \texttt{Scanner}, \texttt{dependency}), (\texttt{hplip}, \texttt{HP Printer}, \texttt{dependency}), (\texttt{xserver-xorg-video-intel}, \texttt{Intel Graphics Card},\\ \texttt{dependency}), etc.), (b) {\em conflict} -- conflict relation refers to a situation where two or more software components, packages, or entities cannot coexist or function harmoniously due to incompatibilities, overlapping functionalities, or conflicting configurations (e.g. triplets include (\texttt{cups}, \texttt{Printers}, \texttt{conflicts}), (\texttt{Keyboard}, \texttt{Input Method Editor (IME)}, \texttt{conflicts}), etc.), {\em affected version} -- these indicate which specific versions of a software are affected by reported issues or bugs (e.g. triplets include (\texttt{Flatbed Scanner}, \texttt{macOS Mojave}, \texttt{affected version}), (\texttt{Error Code 134}, \texttt{sudo dpkg --configure -a}, \texttt{affected version}), etc.), {\em cause and effect} -- these relations refers to cases where one entity (cause) triggers another entity (effect) (e.g. triplets include (\texttt{Error Code 502},~\texttt{nginx},~\texttt{cause and effect)}, (\texttt{Error Code 401}, \texttt{openssh-server}, \texttt{cause and effect}), etc.), and {\em interaction/control} -- corresponds to relations where an entity exerts dynamic influence on / manipulates another entity (e.g. triplets include (\texttt{apt}, \texttt{install}, \texttt{interaction/ control}), (\texttt{docker-ce}, \texttt{run},  \texttt{interaction/control}), etc.).
\noindent\textit{\textbf{Dataset}}: For this experiment we use the earlier 500 bug description data that was manually annotated for the named entities. This data is further annotated with the relationships among the entities by 3 expert annotators with an inter-annotator agreement of 0.693. We end up with a total of 642 triplets, each consisting of a head entity, relationship, and tail entity. Out of these, 27\% have a \textit{dependency} relationship, 21\% are of type \textit{affected versions} while \textit{conflicts}, \textit{cause and effect}, and 
\textit{interaction/control} types account for 4\%, 17\%, and 31\% of the triplets respectively. Since the data for \textit{conflict} type is very less, we ignore it for our experiments.
\noindent\textit{\textbf{Results}}: In Table~\ref{tab:relEx} we show the results of the relation extraction task which is posed as a classification problem having the triplet (the head entity, the tail entity and the relation type) representation as the input and one of the relation classes (\textit{dependency}, \textit{affected versions}, \textit{cause and effect}, and 
\begin{wraptable}{r}{5.5cm}
\vspace*{-0.7cm}
\resizebox{0.43\textwidth}{!}{
\begin{tabular}{|c|c|c|c|}
\hline
\multirow{2}{*}{\textbf{Methods}} &
  \multicolumn{1}{l|}{\multirow{2}{*}{\textbf{BERT-NER}}} &
  \multicolumn{1}{l|}{\multirow{2}{*}{\textbf{BERT-CRF}}} &
  \multicolumn{1}{l|}{\multirow{2}{*}{\textbf{RoBERTa-CRF}}} \\
                                                                      & \multicolumn{1}{l|}{} & \multicolumn{1}{l|}{} & \multicolumn{1}{l|}{} \\ \hline
\textbf{\begin{tabular}[c]{@{}c@{}}Ubuntu\\ (Bug)\end{tabular}}       & \cellcolor{blue!8}0.119                 & \cellcolor{blue!25}0.130                 & 0.101                 \\ \hline
\textbf{\begin{tabular}[c]{@{}c@{}}Launchpad\\ (QA)\end{tabular}} & 0.073                 & \cellcolor{blue!8}0.072                 & \cellcolor{blue!25}0.098                 \\ \hline
\textbf{\begin{tabular}[c]{@{}c@{}}Fedora\\ (CQA)\end{tabular}}       & \cellcolor{blue!8}0.109                 & \cellcolor{blue!25}0.132                 & 0.088                 \\ \hline
\textbf{\begin{tabular}[c]{@{}c@{}}Linux\\ (CQA)\end{tabular}}        & \cellcolor{blue!8}0.122                 & \cellcolor{blue!25}0.148                 & 0.103                 \\ \hline
\end{tabular}
}
\caption{\footnotesize Comparison of performance metrics for BERT-CRF, BERT-NER and RoBERTa-CRF methods across various categories using different datasets (Bug and Launchpad). The best and the second best results are highlighted in \colorbox{blue!25}{dark} and \colorbox{blue!8}{light} purple respectively.}
\label{tab:performance_metrics}
\vspace*{-0.7cm}
\end{wraptable}
\textit{interaction/control}) as output. In the input, we draw the entity representations from our previously trained best NER models (i.e., trained BERT-CRF, trained BERT-NER, and trained RoBERTa-CRF) and compare their performance with vanilla BERT. Naturally, the rationale for employing the trained model as encoder is to acquire a more contextual representation than with the vanilla BERT. Further, we fine-tune the classifier layer with a very small amount of data (3\%) to perform the relation classification. Dark cyan cells in Table~\ref{tab:relEx} represent the best performances and light cyan cells correspond to the second best performances. All the NER based models outperform vanilla BERT for three out of four entity types. Overall, BERT-CRF performs the best.
%In the relation extraction task, we focus on four specific relationships: 'Dependency', 'Affected versions', 'Cause and effect', and 'Interaction/control', and employ four top performing models for our earlier task: BERT-CRF, BERT-NER, RoBERTa-CRF, and Pretrained BERT. The performance of the models varies across different relationships. For the 'Dependency' relationship, BERT-CRF and BERT-NER shows exceptional performance with an F1-score of 0.59, followed by RoBERTa-CRF (0.54). The 'Affected versions' relationship is best identified by BERT-CRF (F1-score 0.15). For the 'Cause and effect' relationship, BERT-CRF and RoBERTa-CRF excel with F1-scores of 0.49, while BERT-NER and Pretrained BERT have second highest score of 0.45. Lastly, for the 'Interaction/control' relationship, Pretrained BERT performs best with the F1-score of 0.48. These results provide a comprehensive understanding of each model's strength in recognizing different types of relationships, which is crucial for guiding future model selection and development. Table~\ref{tab:relEx} displays all the results, using dark cyan to represent the top performance and light cyan to indicate the second highest performance.
\vspace*{-0.3cm}
\section{Error analysis}\label{sec:erroranalysis}
%\SB{Added this section from appendix}
%NER can help extract entities like library names, function names, software versions, bug IDs, or even contributor names from this sea of information. However, it is vital to ensure the NER system's accuracy, and this is where error analysis comes in. 
In this section, we analyze the incorrect predictions and group them into the following types.

\noindent\textit{\textbf{Ambiguity errors}}: Occur when a token's meaning is unclear, such as "Apple" referring to either the OS or an organization. Solutions include enriching training data for ambiguous cases and utilizing advanced models proficient in ambiguity resolution.

\noindent\textit{\textbf{Out-of-Vocabulary (OOV) errors}}: This arises with tokens not in training data, common in the evolving open-source field. Employing character-based or subword-based NER techniques can address OOV issues.

\noindent\textit{\textbf{Boundary detection errors}}: This happens when entity boundaries are wrongly identified, e.g., mistaking "Windows NT3.`.." for separate entities. BIO tagging or expanding training examples with varied entity lengths can help.

\noindent\textit{\textbf{Incorrect entity type errors}}: When entities are correctly identified but misclassified, such as a software version labeled as a date. Broadening the diversity of training examples and refining entity definitions can reduce these errors.

\noindent\textit{\textbf{Cohesion errors}}: Related to software domain semantics, e.g., "sudo, apt, update" should be separate entities. This can be addressed by using sequence tagging models and enriching the training dataset.

\noindent\textit{\textbf{Errors due to homonyms}}: Words with multiple meanings like "Java" need context-aware models to resolve ambiguities based on usage.

\noindent\textit{\textbf{Multi-annotator disagreement}}: Occurs when expert annotators have differing opinions due to the complexity and diversity of software artifacts. Addressing this requires acknowledging and accommodating the range of expert interpretations. %\SB{We can remove if space needed}
\if{0}
\begin{table*}[!h]
\vspace{-0.5cm}
\centering
%\small
\resizebox{1.0\textwidth}{!}{
\begin{tabular}{|l|l|}
\hline
\multicolumn{1}{|c|}{\textbf{Error Type}} & \multicolumn{1}{c|}{\textbf{Samples}}                                                                                                                                                                                                                                                                                                                                                                                                                                                                                                                                                                                                                             \\ \hline
Ambiguity Errors                          & \begin{tabular}[c]{@{}l@{}}('apple', marked as({[}'O', 'B-OS', 'B-Organization'{]})),\\ ('exit',  marked as ({[}'B-Command', 'O', 'B-Software\_Component', 'I-Package'{]})\\ ('ask', marked as ({[}'O', 'B-Package', 'I-Package'{]}))\end{tabular}                                                                                                                                                                                                                                                                                                                                                                                                                \\ \hline
Out-Of-Vocabulary Errors                  & \begin{tabular}[c]{@{}l@{}}dovecot-core → marked as 'O' by human → predicted as 'O' by Bert CRF model → Actually Package\\ Libgtk2.0-bin → marked as Package by Human → marked as Package by Bert CRF model → actually Package\\ Netplan → marked as 'O' by human → predicted as 'O' by Bert CRF model → Actually Package\end{tabular}                                                                                                                                                                                                                                                                                                                            \\ \hline
Boundary Detection Errors                 & \begin{tabular}[c]{@{}l@{}}Windows NT 3.1, Windows NT 3.5, Windows NT 3.51, Windows NT 4.0\\ The model is supposed to start and end till the version.\end{tabular}                                                                                                                                                                                                                                                                                                                                                                                                                                                                                             \\ \hline
Incorrect Entity Type Errors              & \begin{tabular}[c]{@{}l@{}}\textbackslash{}textit\{'I\textbackslash{}'m sure you\textbackslash{}'re aware of the recent "Death by Google Calendar" scandal where somebody had unwittingly\\  published information on Google Calendar that indicated both who they were and in particular, \\ when their house was empty.\}\\ (In the above text “Google” is marked as Package when it should be Organization.)\\ “Nbd-client” is actually a command but is marked as Package by Bert CRF\\“Cursor” is Peripheral but tagged as Software Component.\\“cd” is Command but the model tagged Peripheral (most probably it understood cd rom)\end{tabular} \\ \hline
Cohesion Errors                           & \begin{tabular}[c]{@{}l@{}}For Command - “sudo apt update” our model marks \\ them separately as “sudo”, “apt” \& “update”, but human marked them as a whole.\end{tabular}                                                                                                                                                                                                                                                                                                                                                                                                                                                                                     \\ \hline
\end{tabular}
}
\caption{\footnotesize Error analysis.}
\label{tab:erroranalysis}
\vspace{-0.5cm}
\end{table*}
\fi
\vspace*{-0.2cm}
\section{Conclusion}
In this paper we showed how distant supervision improves the performance of overall NER models in specialized domains like open source softwares where gold labels are scarce. As a follow up step, we also performed the closely-linked task of relation extraction and showed that the NER pipeline improves the extraction performance. In future, we plan to extend this setup for other open source software ecosystems as well as similar specialised domains. %Distant supervision in NER within the open-source domain refers to an approach that utilizes large-scale, pre-labeled data from an unrelated source to improve the effectiveness of the machine learning model in identifying and classifying named entities in text. For example, an open-source NER system could be trained using Wikipedia articles, which are implicitly labeled via hyperlinks and info-boxes. This distant supervision approach can be beneficial when labeled data for a specific domain or task is limited. Similarly, in the field of Relation Extraction (RE), distant supervision allows the application of generalizable patterns from larger datasets to extract relations from unstructured text. The application of distant supervision in RE could enhance the identification of relationships between entities and their attributes, subsequently improving the extraction of complex information from text. It's particularly useful in domains where creating hand-labeled data for every possible relation is impractical due to the vast range of potential relationships.

\bibliographystyle{splncs04}
\bibliography{mybibliography}

\appendix

\section{Error Analysis}
\begin{table*}[!h]
\centering
%\small
\resizebox{1.0\textwidth}{!}{
\begin{tabular}{|l|l|}
\hline
\multicolumn{1}{|c|}{\textbf{Error Type}} & \multicolumn{1}{c|}{\textbf{Samples}}                                                                                                                                                                                                                                                                                                                                                                                                                                                                                                                                                                                                                             \\ \hline
Ambiguity Errors                          & \begin{tabular}[c]{@{}l@{}}('apple', marked as({[}'O', 'B-OS', 'B-Organization'{]})),\\ ('exit',  marked as ({[}'B-Command', 'O', 'B-Software\_Component', 'I-Package'{]})\\ ('ask', marked as ({[}'O', 'B-Package', 'I-Package'{]}))\end{tabular}                                                                                                                                                                                                                                                                                                                                                                                                                \\ \hline
Out-Of-Vocabulary Errors                  & \begin{tabular}[c]{@{}l@{}}dovecot-core → marked as 'O' by human → predicted as 'O' by Bert CRF model → Actually Package\\ Libgtk2.0-bin → marked as Package by Human → marked as Package by Bert CRF model → actually Package\\ Netplan → marked as 'O' by human → predicted as 'O' by Bert CRF model → Actually Package\end{tabular}                                                                                                                                                                                                                                                                                                                            \\ \hline
Boundary Detection Errors                 & \begin{tabular}[c]{@{}l@{}}Windows NT 3.1, Windows NT 3.5, Windows NT 3.51, Windows NT 4.0\\ The model is supposed to start and end till the version.\end{tabular}                                                                                                                                                                                                                                                                                                                                                                                                                                                                                             \\ \hline
Incorrect Entity Type Errors              & \begin{tabular}[c]{@{}l@{}}\textbackslash{}textit\{'I\textbackslash{}'m sure you\textbackslash{}'re aware of the recent "Death by Google Calendar" scandal where somebody had unwittingly\\  published information on Google Calendar that indicated both who they were and in particular, \\ when their house was empty.\}\\ (In the above text “Google” is marked as Package when it should be Organization.)\\ “Nbd-client” is actually a command but is marked as Package by Bert CRF\\“Cursor” is Peripheral but tagged as Software Component.\\“cd” is Command but the model tagged Peripheral (most probably it understood cd rom)\end{tabular} \\ \hline
Cohesion Errors                           & \begin{tabular}[c]{@{}l@{}}For Command - “sudo apt update” our model marks \\ them separately as “sudo”, “apt” \& “update”, but human marked them as a whole.\end{tabular}                                                                                                                                                                                                                                                                                                                                                                                                                                                                                     \\ \hline
\end{tabular}
}
\caption{\footnotesize Error analysis.}
\label{tab:erroranalysis}
\vspace{-0.6cm}
\end{table*}

\section{Prompt}\label{sec:prompt}
\begin{table}[!htb]
\centering
\small
%\resizebox{.45\textwidth}{!}{
\begin{tabular}{|l|}\hline
\textbf{Example Prompt for LLM} \\ \hline
Extract and tag entities along with start and end index and return \\ it in json format from the following paragraph into one of the \\ following entity type: package, operating system, organization,\\ command, error, extension, peripheral, software component,\\ architecture. Paragraph: "...." \\ \hline
\end{tabular}
%}
\caption{\footnotesize Sample prompts to generate entities in LLM.}
\label{tab:promptllm}
\vspace{-0.6cm}
\end{table}
\section{Regex details}\label{sec:regexinfo}
At the beginning of our data processing, we remove any annotations that match directly with common stopwords, as they don't provide meaningful context. When we move to Stage-1 for direct matching, we notice that certain annotations overlap, especially when categorizing specific bugs. To ensure clarity and avoid redundancy, we keep the annotations that cover a more extensive range of data and discard those with shorter overlaps. This means we only use annotations that don't intersect with others. A critical step in our process is to remove all URLs right from the start. We deem entities that overlap with URLs as irrelevant for our analysis. Once URLs are out of the picture, we focus on choosing the remaining annotations that don't overlap, ensuring the integrity and clarity of our data.

\section{Interannotator agreement}
\label{sec:interanno}
Interannotator agreement tends to be somewhat average, and we pinpoint two primary reasons for this:
\begin{compactitem}

   \item 
\noindent\textbf{Ambiguity of entities:} The open-source software domain is expansive and constantly changes. Some terms or phrases might hold different meanings or fall under various categories, depending on the annotator's background and experience in the industry.

\item
\noindent\textbf{Language variability:} In the open-source software sector, language use can be diverse, encompassing technical jargon, acronyms, and even casual speech. This diversity often challenges annotators in consistently recognizing and categorizing entities.

\end{compactitem}
\noindent\textit{Human annotation process:} We engaged four domain experts to undertake our annotation task, dividing them into groups of two for each dataset. All these annotators are majoring in Computer Science contributing
high-quality answers within relevant community platforms (having minimum 3 years of domain knowledge). We divide bugs equally among annotators for each domain and then combine their annotations. They voluntarily joined our project after receiving an invitation through university emails and were rewarded with Amazon gift cards for their contributions. 

\section{Manual labeling in active learning}\label{sec:act-lev-ann}
To classify the entities in the active learning process, we recruited six undergraduate engineering student with expert-level experience in open-source systems and data annotation. Each annotator independently reviewed and classified tagged bugs into one of the nine predefined entity types, requiring a strong understanding of the context and entity characteristics. We provided live training initially to ensure accurate and consistent annotations, which were cross-verified by two researchers. As a token of appreciation and to maintain high motivation, we compensated the annotators with Amazon gift cards.

\section{Hyperparameter settings}\label{sec:appendix3}
Hyperparameter tuning plays a pivotal role in the construction and deployment of any models. The selection of hyperparameters directly impacts an algorithm's learning capacity, with optimal values often resulting in enhanced model performance. In our study, we adopt a methodical strategy for hyperparameter selection. This process starts with grid search, then moves to random search, to efficiently narrow down the feasible range of hyperparameter values. All hyperparameter settings present in Table~\ref{tab:hyperparameter}.
\begin{table*}[!h]
%\tiny
\centering
\resizebox{.75\textwidth}{!}{
\begin{tabular}{|l|l|}
\hline
\multicolumn{1}{|c|}{\textbf{Methods}} & \multicolumn{1}{c|}{\textbf{Hyperparameters}}                                                                                                                                                                                     \\ \hline
Linear CRF                             & "Passive Aggressive" algorithm, 150 iterations                                                                                                                                                                                         \\ \hline
BiLSTM CRF                             & \begin{tabular}[c]{@{}l@{}}embedding dim= 768, BiLSTM dim = 256, LEARNING\_WEIGHT = 5e-2 \\ WEIGHT\_DECAY = 1e-4, epochs = 3\end{tabular}                                                                                              \\ \hline
Bert NER                               & \begin{tabular}[c]{@{}l@{}}dropout = 0.1, max\_seq = 512, AdamW, epochs = 15, lr: 5.0e-06, batch\_size = 32\\ lr\_scheduler:\\     end\_factor: 0.0,start\_factor: 1.0,total\_iters: 25, type: LinearLR\end{tabular}  \\ \hline
Bert CRF                               & \begin{tabular}[c]{@{}l@{}}dropout = 0.1, max\_seq = 512, AdamW, epochs = 15, lr: 5.0e-06, batch\_size = 32\\ lr\_scheduler:\\     end\_factor: 0.0,start\_factor: 1.0,total\_iters: 25,type: LinearLR\end{tabular}  \\ \hline
Partial Bert CRF                       & \begin{tabular}[c]{@{}l@{}}dropout = 0.1, max\_seq = 512, AdamW, epochs = 15, lr: 5.0e-06, batch\_size = 32\\ lr\_scheduler:\\     end\_factor: 0.0,start\_factor: 1.0,total\_iters: 25,type: LinearLR\end{tabular}  \\ \hline
spanBert CRF                           & \begin{tabular}[c]{@{}l@{}}dropout = 0.1, max\_seq = 512, AdamW, epochs = 5, lr: 5.0e-06, batch\_size = 11, \\ lr\_scheduler:\\     end\_factor: 0.0,start\_factor: 1.0,total\_iters: 30,type: LinearLR\end{tabular} \\ \hline
SoftNER                                & epochs = 10, bert-base-uncased, max\_seq = 512, lr = 5e-5,epsilon for adam optimiser = 1e-8                                                                                                                                            \\ \hline
Roberta CRF                            & \begin{tabular}[c]{@{}l@{}}dropout = 0.1, max\_seq = 512, AdamW, epochs = 15, lr: 5.0e-06, batch\_size = 32\\ lr\_scheduler:\\     end\_factor: 0.0,start\_factor: 1.0,total\_iters: 25,type: LinearLR\end{tabular}  \\ \hline
Pretrained Roberta CRF                 & \begin{tabular}[c]{@{}l@{}}dropout = 0.1, max\_seq = 512, AdamW, epochs = 5, lr: 5.0e-06, batch\_size = 32\\ lr\_scheduler:\\     end\_factor: 0.0,start\_factor: 1.0,total\_iters: 25,type: LinearLR\end{tabular}   \\ \hline
\end{tabular}
}
\caption{\footnotesize Hyperparameters.}
\label{tab:hyperparameter}
\end{table*}

%\begin{longtable*}[ht]
\begin{table*}[ht]
\centering
\resizebox{.99\textwidth}{!}{
\begin{tabular}{|c|c|c|}
\hline
\textbf{LLM Models} & \textbf{Annotated Data} & \textbf{Index} \\ \hline
GPT-3.5-Turbo       & \includegraphics[scale = 0.5]{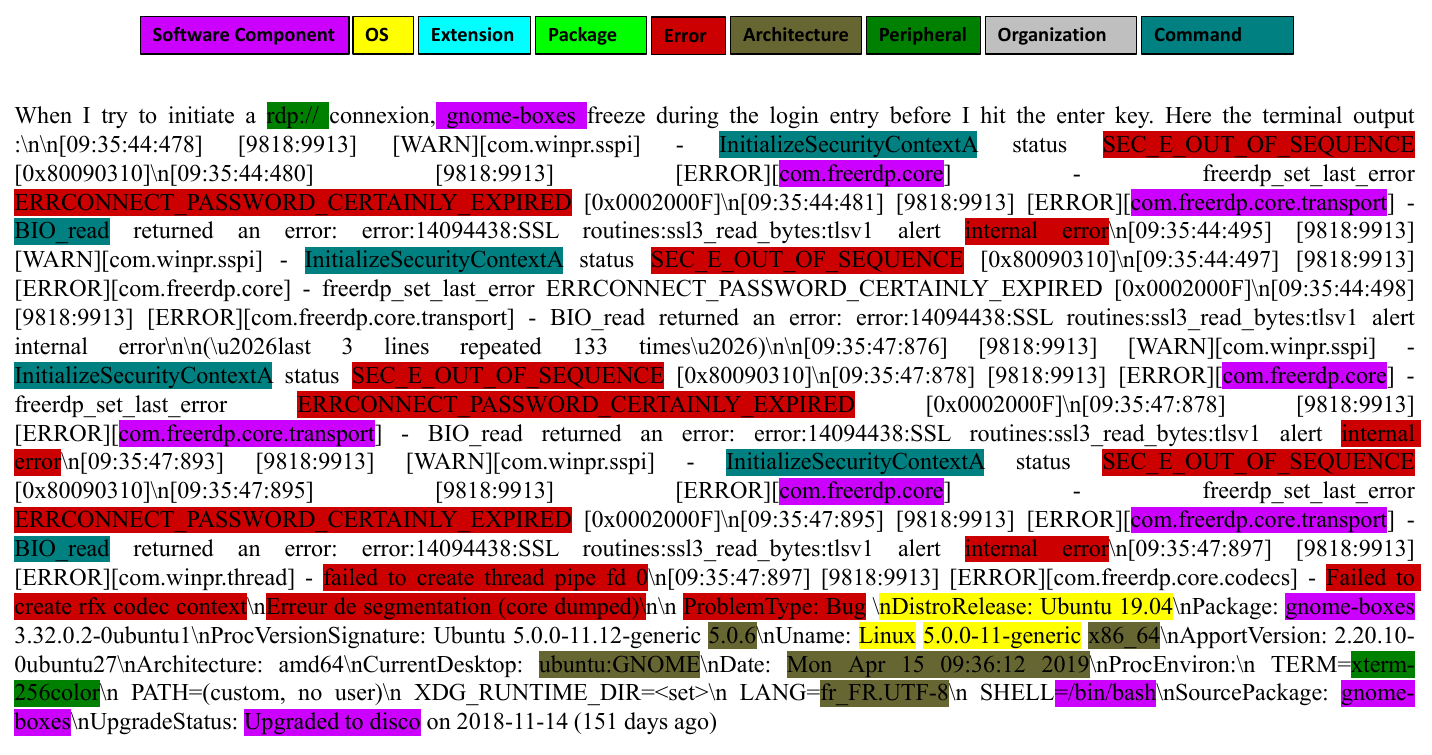}                       & 1              \\ \hline
GPT-4               &  \includegraphics[scale = 0.5]{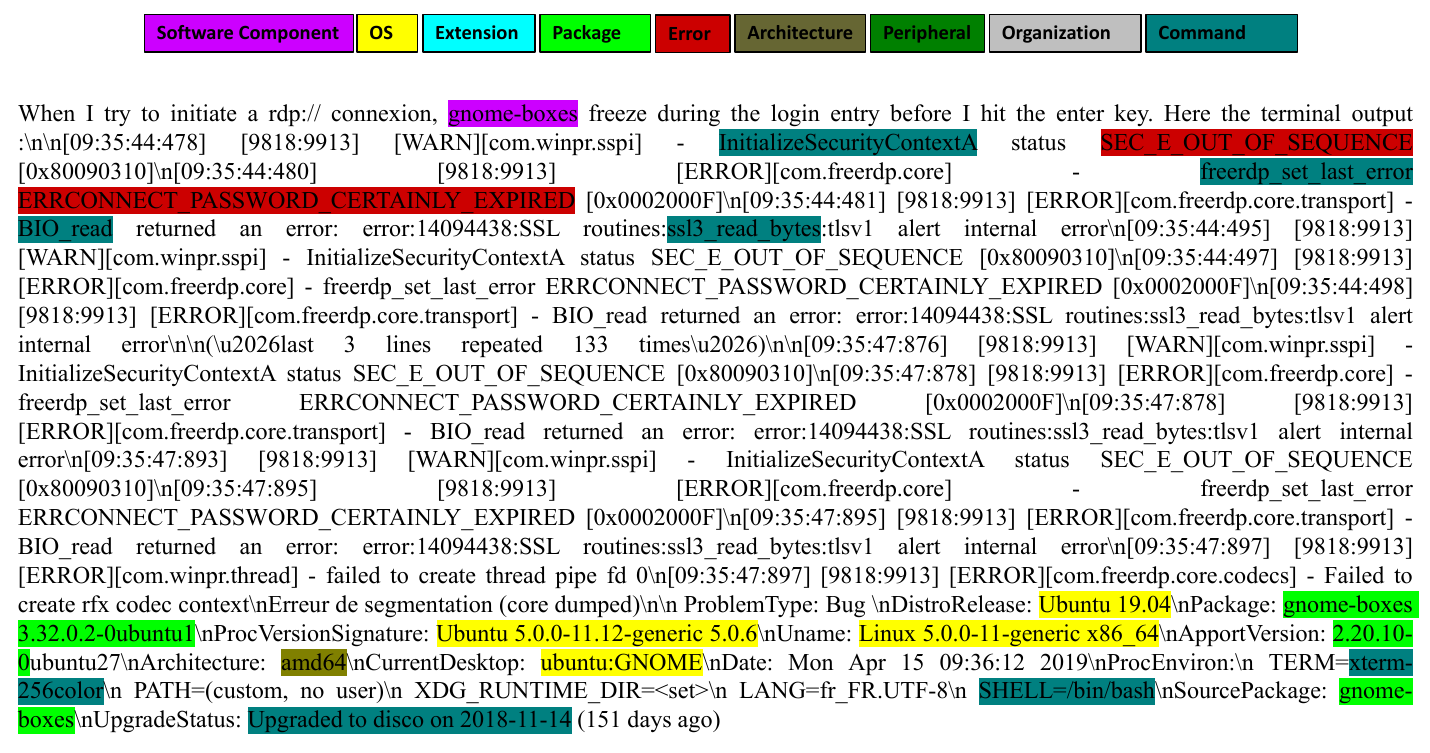}                       & 1              \\ \hline
Google BARD         & \includegraphics[scale = 0.5]{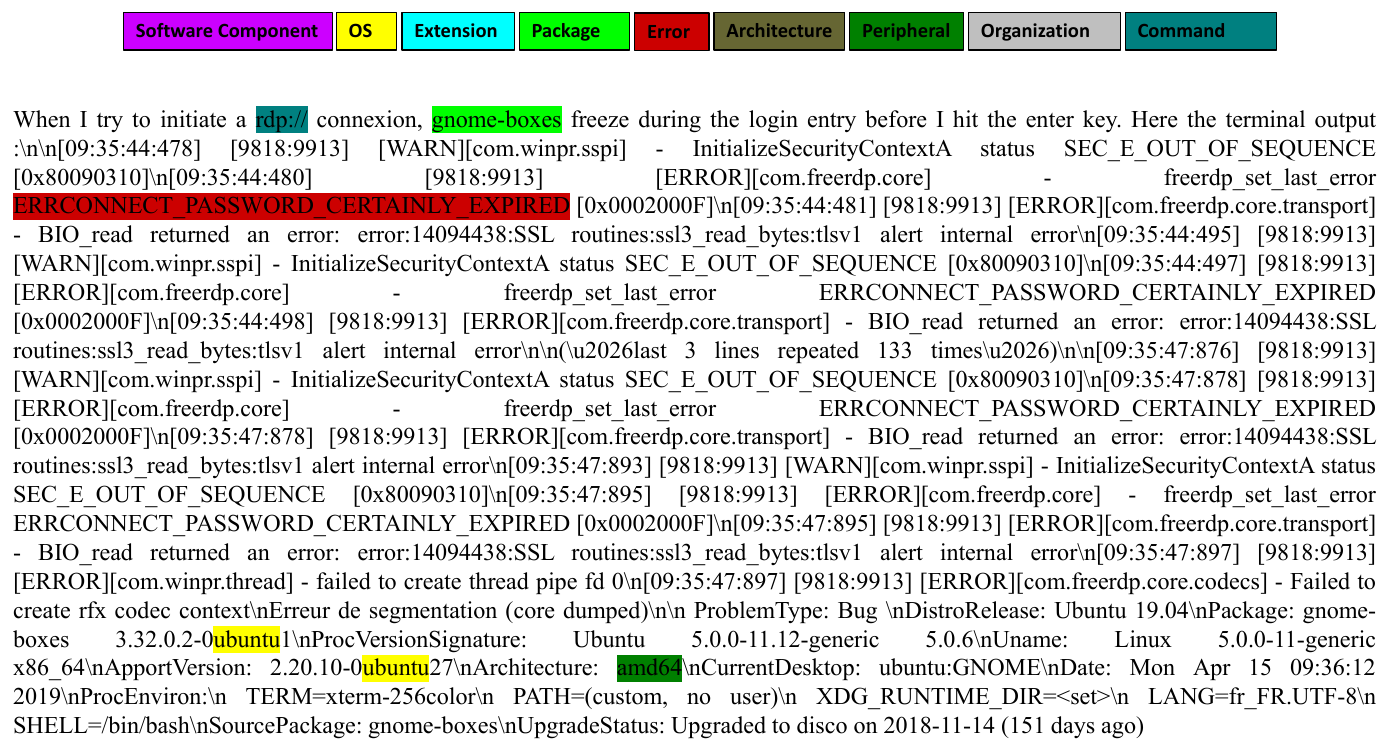}                        & 1              \\ \hline
\end{tabular}
}
\caption{\footnotesize Sample LLM predictions.}
\label{tab:llmprediction1}
\end{table*}
%\end{longtable*}

\begin{table*}[ht]
\centering
\resizebox{.99\textwidth}{!}{
\begin{tabular}{|c|c|c|}
\hline
\textbf{LLM Models} & \textbf{Annotated Data} & 
\textbf{Index} \\ \hline
GPT-3.5-Turbo       & \includegraphics[scale = 0.5]{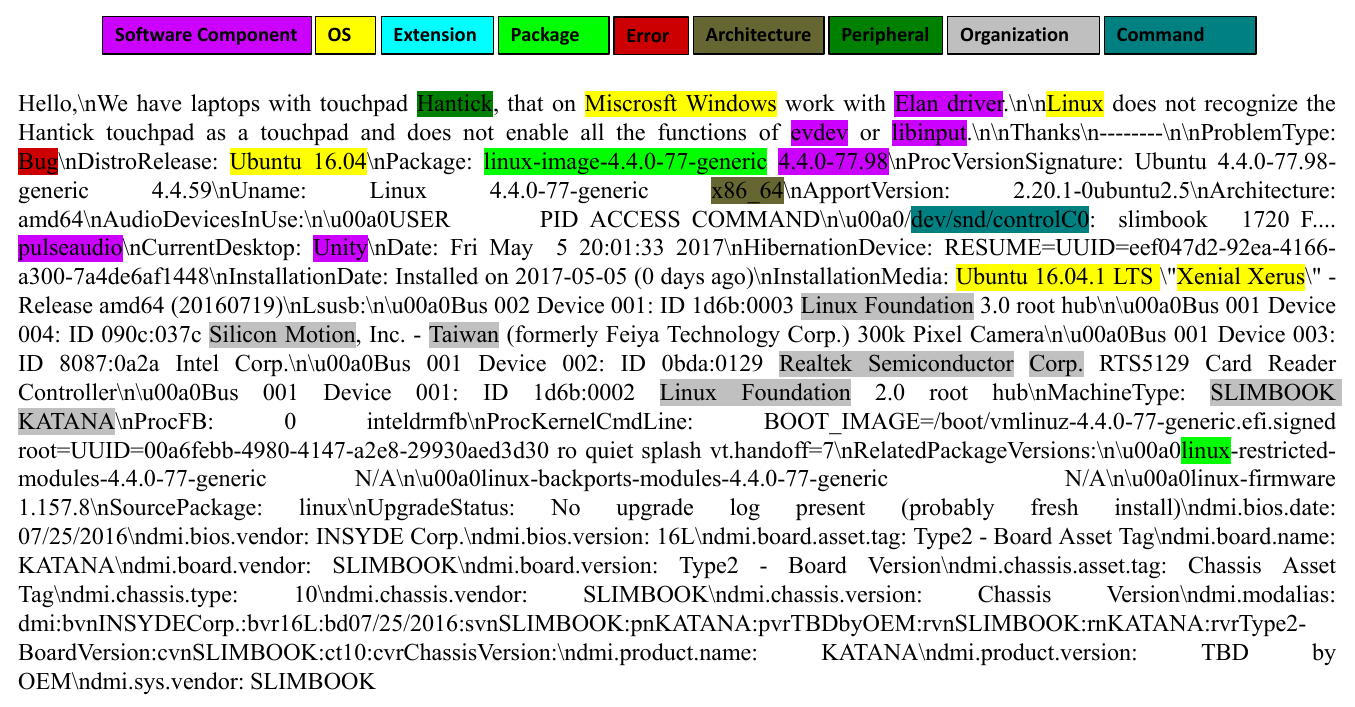}                        & 2              \\ \hline
GPT-4               & \includegraphics[scale = 0.5]{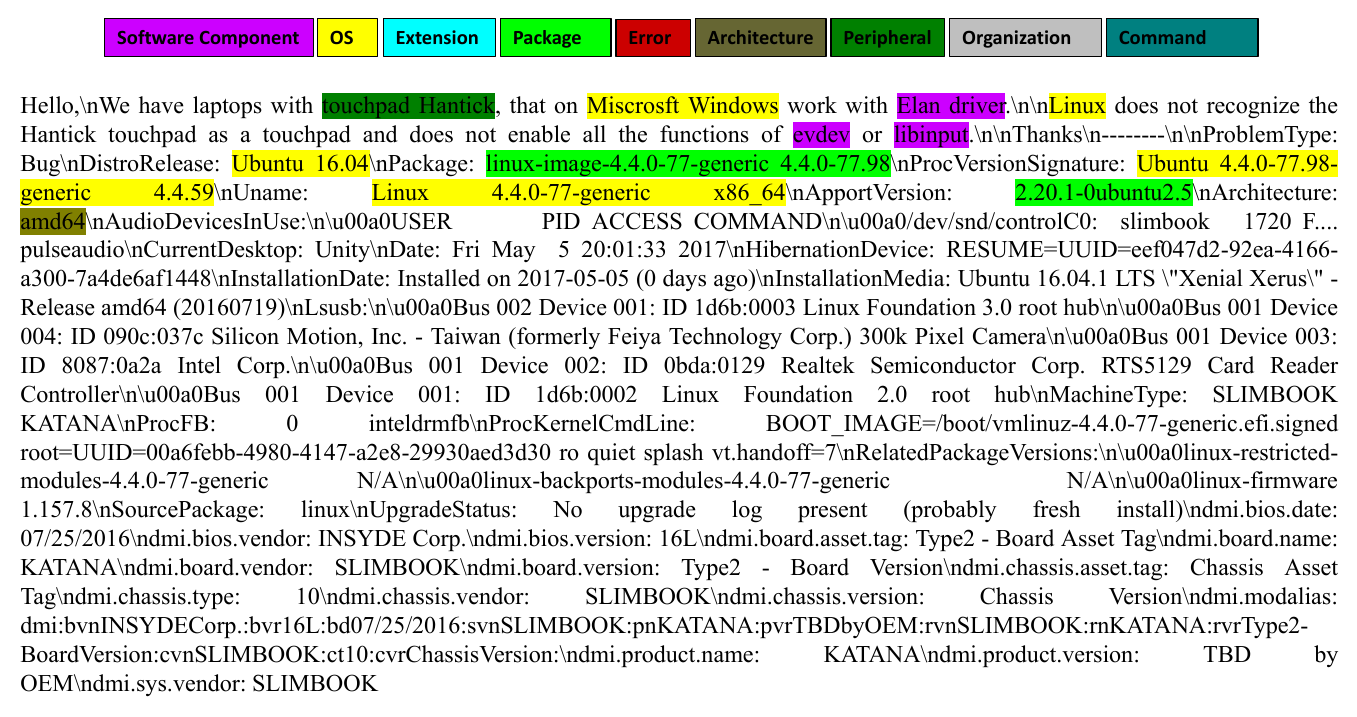}                        & 2              \\ \hline
Google BARD         & \includegraphics[scale = 0.5]{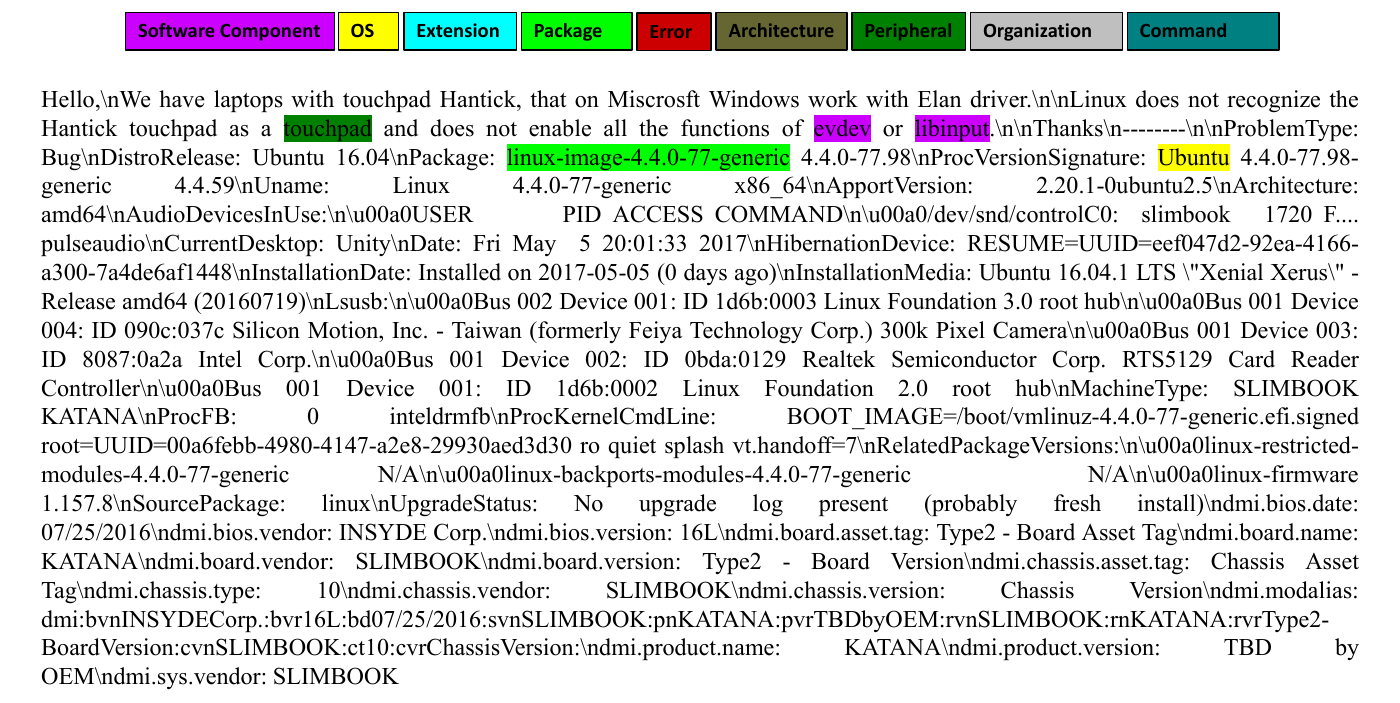}                        & 2              \\ \hline
\end{tabular}
}
\caption{\footnotesize Sample LLM predictions.}
\label{tab:llmprediction2}
\end{table*}
\end{document}